\documentclass[preprint,12pt]{elsarticle}

\usepackage[T1]{fontenc}
\usepackage[margin=0.9in]{geometry}
\usepackage{graphics}
\usepackage{amsmath,amssymb,amsfonts}%
\usepackage{amsthm}%
\usepackage{mathrsfs}%
\usepackage[title]{appendix}%
\usepackage{xcolor}%
\usepackage{textcomp}%
\usepackage{manyfoot}%
\usepackage{booktabs}%
\usepackage{algorithm}%
\usepackage{algorithmicx}%
\usepackage{algpseudocode}%
\usepackage{listings}%
\usepackage{indentfirst}
\usepackage{lineno}
\usepackage{mathrsfs}
\usepackage{array}
\usepackage{float}
\usepackage{verbatim}
\usepackage{setspace}
\usepackage{makecell}
\usepackage{textcomp}
\usepackage{multirow}%

\usepackage{bm}
\usepackage{url}
\usepackage{hyperref}
\usepackage{marvosym}
\usepackage{cleveref}

\usepackage{amsmath,amssymb,amsfonts}
\usepackage{amsthm}
\usepackage{mathrsfs}
\usepackage{wasysym}
\usepackage{enumitem}
\usepackage{bm}
\usepackage{makecell}
\usepackage{diagbox}

\usepackage{multirow,booktabs,comment}
\usepackage{epstopdf}
\usepackage{wrapfig}
\usepackage{subfig}

\usepackage[normalem]{ulem}

\usepackage{soul}

\newtheorem*{remark}{Remark}

\newcommand{\bx}{\boldsymbol{x}}

\newcommand{\norm}[2]{\left\| #1 \right\|_{#2}}

\newcommand{\mb}{\boldsymbol}

\newcommand{\xs}{\mathbb}

\newcommand{\comm}[1]{}

\makeatletter
\def\ps@pprintTitle{%
   \let\@oddhead\@empty
   \let\@evenhead\@empty
   \let\@oddfoot\@empty
   \let\@evenfoot\@oddfoot
}
\makeatother

\begin{document}

\begin{frontmatter}



\title{Estimating Committor Functions via Deep Adaptive Sampling on Rare Transition Paths}

 \author[mymainaddress]{Yueyang Wang}
\ead{wangyueyang@stu.pku.edu.cn}

\author[myfirstaddress]{Kejun Tang\textsuperscript{\Letter}}
\ead{tangkejun@suat-sz.edu.cn}

 \author[mymainaddress]{Xili Wang}
\ead{xiliwang@stu.pku.edu.cn}

 \author[mysecondaddress]{Xiaoliang Wan}
\ead{xlwan@lsu.edu}

 \author[mythirdaddress]{Weiqing Ren}
\ead{matrw@nus.edu.sg}

 \author[mymainaddress]{Chao Yang}
	\ead{chao\_yang@pku.edu.cn}

 \address[mymainaddress]{School of Mathematical Sciences, Peking University}
 \address[myfirstaddress]{Faculty of Computility Microelectronics, Shenzhen University of Advanced Technology}
 \address[mysecondaddress]{Department of Mathematics and Center for Computation and Technology, Louisiana State University}
  \address[mythirdaddress]{Department of Mathematics, National University of Singapore}

\begin{abstract}
		The committor functions are central to investigating rare but important events in molecular simulations. It is known that computing the committor function suffers from the curse of dimensionality. Recently, using neural networks to estimate the committor function has gained attention due to its potential for high-dimensional problems. Training neural networks to approximate the committor function needs to sample transition data from straightforward simulations of rare events, which is very inefficient. The scarcity of transition data makes it challenging to approximate the committor function. To address this problem, we propose an efficient framework to generate data points in the transition state region that helps train neural networks to approximate the committor function. We design a Deep Adaptive Sampling method for TRansition paths (DASTR), where deep generative models are employed to generate samples to capture the information of transitions effectively. In particular, we treat a non-negative function in the integrand of the loss functional as an unnormalized probability density function and approximate it with the deep generative model. The new samples from the deep generative model are located in the transition state region and fewer samples are located in the other region. This distribution provides effective samples for approximating the committor function and significantly improves the accuracy. We demonstrate the effectiveness of the proposed method through both simulations and realistic examples. 
\end{abstract}




\end{frontmatter}




\section{Introduction}
Understanding transition events between metastates in a stochastic system plays a central role in chemical
reactions and statistical physics \citep{okuyama1998transition, E2006towards, berteotti2009protein,E2010transition}. The physical process can be formulated as the following stochastic differential equation (SDE)
\begin{equation}\label{eq_committor_sde}
	d \mb{X}_t = -\nabla{V(\mb{X}_t)}dt + \sqrt{2 \beta^{-1}} d\mb{W}_t,
\end{equation}
where $\mb{X}_t \in \Omega \subset \xs{R}^d$ is the state of the system at time $t$, $V: \Omega \mapsto \xs{R}$ denotes a potential function, $\beta$ is the inverse temperature, and $\mb{W}_t$ is the standared $d$-dimensional Wiener process. For two disjoint subsets of this stochastic system, we are interested in the transition rate, which can be characterized by the \emph{commttor function}. For two distinct metastable regions $A, B \subset \Omega$, and $A \cap B = \emptyset$, denoting by $\tau_{\omega}$ the first hitting time of a subset $\omega \subset \Omega$ for a trajectory, the committor function $q: \Omega \mapsto [0,1]$ is defined as $q(\mb{x}) = \xs{P} \left( \tau_{B} < \tau_{A} |\mb{X}_0 = \mb{x}\right)$,
where $\xs{P}$ denotes the probability. The committor function is a probability that a trajectory of SDE starting from $\mb{x} \in \Omega$ first reaches $B$ rather than $A$. By definition, it is easy to verify that $q(\mb{x}) = 0$ for $\mb{x} \in A$ and $q(\mb{x}) = 1$ for $\mb{x} \in B$.
This committor function provides the information of process of a transition, and it is governed by the following partial differential equation (PDE) \cite{lai2018point,li2019computing}
\begin{equation}\label{eq_committor_pde}
	\begin{aligned}
		-\beta^{-1} \Delta q(\mb{x}) + \nabla V(\mb{x}) \cdot \nabla q(\mb{x}) &= 0, \quad \mb{x} \in \Omega \backslash (A \cup B), \\
		q(\mb{x}) &= 0, \quad \mb{x} \in A, \\
		q(\mb{x}) &=1 , \quad \mb{x} \in B, \\
		\nabla q(\mb{x}) \cdot \mb{n} &= 0, \quad \mb{x} \in \partial \Omega \backslash (A \cup B),
	\end{aligned}
\end{equation}
where $\mb{n}$ is the outward unit normal vector of the boundary $\partial \Omega \backslash (A \cup B)$. Once the committor function $q(\mb{x})$ is found, we can use it to extract the statistical information of reaction trajectories \citep{E2006towards,E2010transition}.

\subsection{Connections with Prior Work and Contributions}
Obtaining the committor function $q$ needs to solve the above high-dimensional PDE, which is computationally infeasible for traditional grid-based numerical methods. In \cite{chen2023committor}, a low-rank tensor train approach is proposed to compute the committor function, which relies on the low-rank tensor train approximation of the Boltzmann-Gibbs distribution. This approach cannot be directly applied to realistic problems if no explicit low-rank tensor train formats for the potential are given.
Some efforts have been made to employ deep neural networks to approximate the committor function \citep{khoo2019solving,li2019computing,li2022semigroup}. The key idea is that committor functions are represented by deep neural networks that can be trained by minimizing a variational loss functional. The training data points for discretizing the variational loss are usually sampled from the equilibrium distribution of the SDE (i.e. the Gibbs measure) \citep{khoo2019solving, li2022semigroup, li2020solving}, which needs to simulate the stochastic differential equations. This sampling method is inefficient due to the scarcity of transition data, especially for realistic systems at low temperatures.   
Modified sampling methods are proposed in \citep{li2019computing,rotskoff2022active,hasyim2022supervised,kang2024computing,lin2024deep} to alleviate this issue, where a new probability measure for sampling is constructed by modifying the potential function so that more samples can be obtained in the transition state region.

When the transition is rare, samples from the transition state region are difficult to obtain from the SDE \citep{rotskoff2022active,kang2024computing}. 
If insufficient data points are located on the transition paths, the trained neural network for approximating the committor function will have a large generalization error. To address this problem, we propose a new framework called Deep Adaptive Sampling on rare TRansition paths (DASTR) to train the deep neural network. More specifically, we generate samples in the transition state region using an iterative construction. To do this, we define a proper sampling distribution using both the current approximate committor function and the potential function, in contrast to merely modifying the potential function as in \cite{li2019computing,rotskoff2022active,hasyim2022supervised,kang2024computing,lin2024deep}. The key idea is to reveal the transition information by taking into account the properties of the committor function. The new distribution is approximated by a deep generative model based on which new samples are generated and added to the training set. Once the training set is updated, the neural network model for approximating the committer function is further trained for refinement. This procedure is repeated to form a deep adaptive sampling approach on rare transition paths.

It is challenging to deal with high-dimensional realistic problems using deep generative models because we need to ensure two things: one is that more samples are located in the transition state region, and the other is that all samples must obey the molecular configurations. Directly approximating and sampling a high-dimensional distribution may result in a relatively large number of samples with unreasonable molecular configurations, which limits the application of DASTR. 
To deal with this issue, we combine the proposed DASTR method with dimension reduction techniques to automatically select the collective variables (CVs), where an autoencoder is trained to help avoid hand-craft selections of collective variables. Such a dimension reduction step helps avoid generating physically unreasonable configurations, thereby not only reducing computational complexity but also  enhancing sampling efficiency.
To summarize, the main contributions of this work are as follows:
\begin{itemize}
	\item We propose a general framework, called deep adaptive sampling on rare transition paths (DASTR), for estimating high-dimensional committor functions. 
	\item For high-dimensional realistic problems, the proposed DASTR method can be applied to the latent collective variables obtained by an autoencoder without hand-picking. One can reduce computational complexity and enhance sampling efficiency by adaptive sampling in the latent space. We demonstrate the effectiveness of the proposed method with the alanine dipeptide problem.
\end{itemize}

\subsection{Related Work}

\paragraph{Adaptive Sampling of Neural Network Solver} The basic idea of adaptive sampling involves utilizing a non-negative error indicator, such as the residual square, to refine collocation points in the training set. Sampling approaches \citep{gao2021active} (e.g., Markov Chain Monte Carlo) or deep generative models \citep{tang_das,wang_das2,tang_aas} are often invoked to sample from the distribution induced by the error indicator. Typically, an additional deep generative model (e.g., normalizing flow models) or a classical model (e.g., Gaussian mixture models \citep{gao2023failure, jiao2023gas}) for sampling is required. This work uses the variational formulation and defines a novel indicator for adaptive sampling by incorporating the traits of committor functions. 

\paragraph{Autoencoder for Protein Systems} 
As a dimension reduction technique, autoencoders have shown the potential for the protein structure prediction and generation \citep{czibula2021autoppi}. Autoencoders compress the input data into a lower-dimensional latent space and then reconstruct the input data through a decoder, enabling the learning of underlying features in the data. This approach not only helps reduce the computational resources needed for protein simulations but also significantly lowers the dimensionality and complexity of the problem. The prediction and generation of new protein structures can also be assisted by analyzing the variables in the latent space \citep{alam2019learning, hawkins2021generating}. In our framework, the deep generative model can be used in the latent space to adaptively generate latent variables, which helps us explore the transition paths more efficiently and avoid selecting collective variables by hand-picking.

\subsection{Organization}
The rest of the paper is organized as follows. Details of neural network methods for computing committor functions are introduced in section \ref{sec_nn_committor}. Our DASTR approach is presented in section \ref{sec_dastr}. In section \ref{sec_numexp}, we demonstrate the effectiveness of our DASTR approach with numerical experiments. Finally, we make a conclusion in section \ref{sec_concl}.

\section{Neural Network Solver for Committor Functions}\label{sec_nn_committor}
The neural network approximation of partial differential equations involves minimizing a proper loss functional, e.g., the residual loss \citep{sirignano2018dgm,raissi2019physics,karniadakis2021physics} or the variational loss \citep{weinan2018deep,LMDeepNitscheMethod,lu2021priori}. For the committor function, we consider the variational loss \citep{li2019computing} instead of the residual loss. The variational loss involves up to first-order derivatives of the committer function,  while the residual loss needs to compute the second-order derivatives. In other words, computing the residual loss is more expensive, especially for high dimensional problems (large $d$ in \eqref{eq_committor_pde}).
Let $q_{\mb{\theta}}(\mb{x})$ be a neural network parameterized with $\mb{\theta}$, where the input of the neural network is the state variable $\mb{x}$. One can solve the following variational problem to approximate the committor function
\begin{equation}\label{eq_committor_var}
	\begin{aligned}
		& \ \min_{\mb{\theta}}  \int_{\Omega \backslash (A \cup B)} \vert \nabla q_{\mb{\theta}}(\mb{x}) \vert^2 e^{-\beta V(\mb{x})} d\mb{x}, \\
		& \text{s.t.}  \ q_{\mb{\theta}}(\mb{x}) = 0, \mb{x} \in A; q_{\mb{\theta}}(\mb{x}) = 1, \mb{x} \in  B. 
	\end{aligned}
\end{equation}
The details of the derivation of \eqref{eq_committor_var} can be found in \ref{sec_appendix_var}. We then obtain the following unconstrained optimization problem by adding a penalty term
\begin{equation} \label{eq_committor_uncons}
	\min_{\mb{\theta}} \int_{\Omega \backslash (A \cup B)} \vert \nabla q_{\mb{\theta}}(\mb{x}) \vert^2 e^{-\beta V(\mb{x})} d\mb{x} + \lambda\bigg(\int_A  q_{\mb{\theta}}(\mb{x})^2 p_A(\mb{x}) d\mb{x} + \int_B  (1-q_{\mb{\theta}}(\mb{x}))^2p_B(\mb{x})d\mb{x}\bigg),
\end{equation}
where $\lambda > 0$ is a penalty parameter, $p_A$ and $p_B$ are two probability density functions on $A$ and $B$ respectively.

To optimize the above variational problem, one needs to generate some random collocation points from a proper probability distribution to estimate the integral in \eqref{eq_committor_var}. 
One choice is to sample collocation points from the Gibbs measure $e^{-\beta V(\mb{x})}/Z$, where $Z = \int_{\Omega \backslash (A \cup B)} e^{-\beta V(\mb{x})} d\mb{x} $ is the normalization constant, and this can be done by simulating the SDE defined in \eqref{eq_committor_sde}. However, generating collocation points by the SDE is inefficient for approximating the committor function, especially for molecular systems with low temperatures (or high energy barriers). This is because the committor function focuses on the transition area while the samples generated by the Langevin dynamics (equation \eqref{eq_committor_sde}) cluster around the metastable regions $A$ and $B$. This implies that the samples from the SDE may not include sufficient effective samples for training $q_{\mb{\theta}}$. Hence, we need a strategy to seek more effective samples to approximate the committor function, which will be presented in the next section. 

Now suppose that we have a set of collocation points $\mathsf{S} = \{\mb{x}_{i} \}_{i=1}^N$, where each $\mb{x}_{i} \in \Omega \backslash (A \cup B)$ is drawn from a certain probability distribution $p$, and two sets of collocation points $\mathsf{S}_A = \{\mb{x}_{A, i} \}_{i=1}^{N_A}$ and $\mathsf{S}_B = \{\mb{x}_{B, i} \}_{i=1}^{N_B}$, where each $\mb{x}_{A,i}$ and each $\mb{x}_{B,i}$ are drawn from $p_{A}$ and $p_{B}$ respectively. The optimization problem \eqref{eq_committor_uncons} can be discretized as follows
\begin{equation} \label{eq_vardiscrete}
	\min_{\mb{\theta}} \frac{1}{N} \sum\limits_{i=1}^N  \vert \nabla q_{\mb{\theta}}(\mb{x}_{i}) \vert^2 \frac{e^{-\beta V(\mb{x}_i)}}{p(\mb{x}_i)} + \lambda  \left(\frac1{N_A}\sum_{i=1}^{N_A}q_{\mb{\theta}}(\mb{x}_{A,i})^2 + \frac1{N_B}\sum_{i=1}^{N_B}(q_{\mb{\theta}}(\mb{x}_{B,i})-1)^2\right).
\end{equation}
The key point here is to choose an effective set $\mathsf{S}$ to train $q_{\mb{\theta}}$. In the next section, we will show how to adaptively generate effective collocation points (a high-quality dataset) on rare transition paths, based on which we can improve the accuracy of the approximate solution of \eqref{eq_committor_pde}. Considering that the main difficulties come from the transition state region, we will focus on how to choose $\mathsf{S}$ and assume that the integral on the boundary is well approximated by two prescribed sets $\mathsf{S}_A$ and $\mathsf{S}_B$. For simplicity, we will ignore the penalty term when discussing our method.

\section{Deep Adaptive Sampling on Rare Transition Paths}\label{sec_dastr}

\subsection{Main Idea}
Our goal is to adaptively generate more effective data points distributed in the region of the transition state. This will be  achieved by designing a deep adaptive sampling method on the transition paths.

Suppose that at the $k$-th step, we have obtained the current approximate solution $q_{\mb{\theta}_k}$ with $\mathsf{S}_k$. We want to use the information of $q_{\mb{\theta}_k}$ and the potential function $V$ to detect where the transition area is, based on which we expect to generate new data points in the transition state region that can help improve the discretization given by $\mathsf{S}_{k}$. We  then refine $\mathsf{S}_{k}$ to get $\mathsf{S}_{k+1}$ for the next training step. The more effective data points in the transition area we have, the more accurate solution $q_{\mb{\theta}}$ we can obtain. To achieve this, we define a proper probability distribution for sample generation based on the following observations: First, 
$|\nabla_{\mb{x}} q|^2$ has a peak in the transition state region, implying that more data points should be introduced around the peak. Second, we may lower the energy barrier to facilitate transitions between the metastable states, which can be done by adding a biased potential $V_{\text{bias}}$ to the original potential $V$ \citep{li2019computing,kang2024computing}. 

\subsection{Sample Generation}\label{sec_dastr_sample}
Let $p_{V, q}$ be a probability density function (PDF) that is dependent on $V$ and $q_{\mb{\theta}}$. Here, we give two choices for constructing $p_{V, q}$. One choice is to set 
\begin{equation}\label{eq_committor_pdf1}
	p_{V, q}(\mb{x}) = \frac{\vert \nabla q_{\mb{\theta}}(\mb{x}) \vert^2 e^{-\beta V(\mb{x})}}{C_1},
\end{equation}
where $C_1$ is the normalization constant. That is, we treat the nonnegative integrand in \eqref{eq_committor_var} as an unnormalized probability density function. If there exists a high energy barrier, we can use a biased potential $V_{\text{bias}}$ to lower the energy barrier, which yields the following sampling distribution 
\begin{equation}\label{eq_committor_pdf2}
	p_{V, q}(\mb{x}) = \frac{\vert \nabla_{\mb{x}} q_{\mb{\theta}}(\mb{x}) \vert^2 e^{-\beta (V(\mb{x}) + V_{\text{bias}}(\mb{x}))}}{C_2},
\end{equation}
where $C_2$ is the corresponding normalization constant. The biased potential can be chosen to be an umbrella potential \citep{kastner2011umbrella} or a potential derived from the metadynamics \citep{bussi2020using, barducci2008well}.

Now the question is how we can generate samples from the above sampling distribution. Here, we use KRnet, which is a type of flow-based generative models \citep{dinh2016density,kingma2018glow}, for PDF approximation and sample generation. We note that other deep generative models with exact likelihood computation \citep{chen2018neural, song2021scorebased} can also be used here. Let $p_{\mathsf{KRnet}}(\mb{x};\Theta_f)$ be a PDF model induced by KRnet with parameters $\Theta_f$ \citep{tang_das, tangwandensity2020,wan2022vae, tang2022adaptive}. The PDF model $p_{\mathsf{KRnet}}$ is induced by a bijection $f_{\mathsf{KRnet}}$ with parameters $\Theta_f$:
\begin{equation*}\label{eq_krpdf}
	p_{\mathsf{KRnet}}(\mb{x};\Theta_f)=p_{\mb{Z}}(f_{\mathsf{KRnet}}(\mb{x})) \left |\det\nabla_{\mb{x}} f_{\mathsf{KRnet}} \right|,
\end{equation*} 
where $p_{\mb{Z}}$ is a prior PDF (e.g., the standard Gaussian distribution). We can approximate $p_{V, q}$ through solving the optimization problem
\begin{equation*}\label{eqn:KL_opt}
	\Theta_f^*=\arg \min_{\Theta_f} D_{\mathsf{KL}}(p_{V, q}(\mb{x}) \| p_{\mathsf{KRnet}}(\bx;\Theta_f)),
\end{equation*}
where $D_{\mathsf{KL}}(\cdot\|\cdot)$ indicates the Kullback-Leibler (KL) divergence between two distributions. 
Minimizing the KL divergence is equivalent to minimizing the cross entropy between $p_{V, q}$ and $p_{\mathsf{KRnet}}$ \citep{de2005tutorial, rubinstein2013cross}: 
$$H(p_{V, q}, p_{\mathsf{KRnet}})=- \int_{\Omega \backslash (A \cup B)}p_{V, q}(\mb{x})  \log p_{\mathsf{KRnet}}(\bx;\Theta_f) d\mb{x}.$$
The normalization constants in \eqref{eq_committor_pdf1} and \eqref{eq_committor_pdf2} do not affect the optimization with respect to $\Theta_f$. 
Since the samples from $p_{V, q}$ are not available, one can approximate the cross entropy using the importance sampling technique:
\begin{equation}\label{eqn:ce_approx}
	H(p_{V, q}, p_{\mathsf{KRnet}}) \approx -\frac{1}{N}\sum_{i=1}^{N} \frac{p_{V, q}(\mb{x}_i)}{p_{\mathsf{IS}}(\mb{x}_i)}\log p_{\mathsf{KRnet}}(\mb{x}_i;\Theta_f),
\end{equation}
where $p_{\mathsf{IS}}(\mb{x}_i)$ is a known PDF model and $\{\mb{x}_i\}_{i=1}^{N}$ are the samples from $p_{\mathsf{IS}}(\mb{x}_i)$. For example, the PDF model $p_{\mathsf{IS}}(\mb{x}_i)$ can be chosen to be a PDF model induced by a known KRnet with parameters $\Theta_f^{\prime}$, i.e.,
\begin{equation}\label{eqn:z_to_x}
	\mb{x}_i = f^{-1}_{\mathsf{KRnet}}(\mb{z}_i),
\end{equation}
with $\mb{z}_i$ being sampled from the standard Gaussian distribution. 
We then minimize the discretized cross entropy \eqref{eqn:ce_approx} to obtain an approximation of $\Theta_f^*$.

\subsection{DASTR Algorithm}
\begin{figure}[!htb]
	\begin{center}
		\vspace{-1em}
		\includegraphics[width=.86\textwidth]{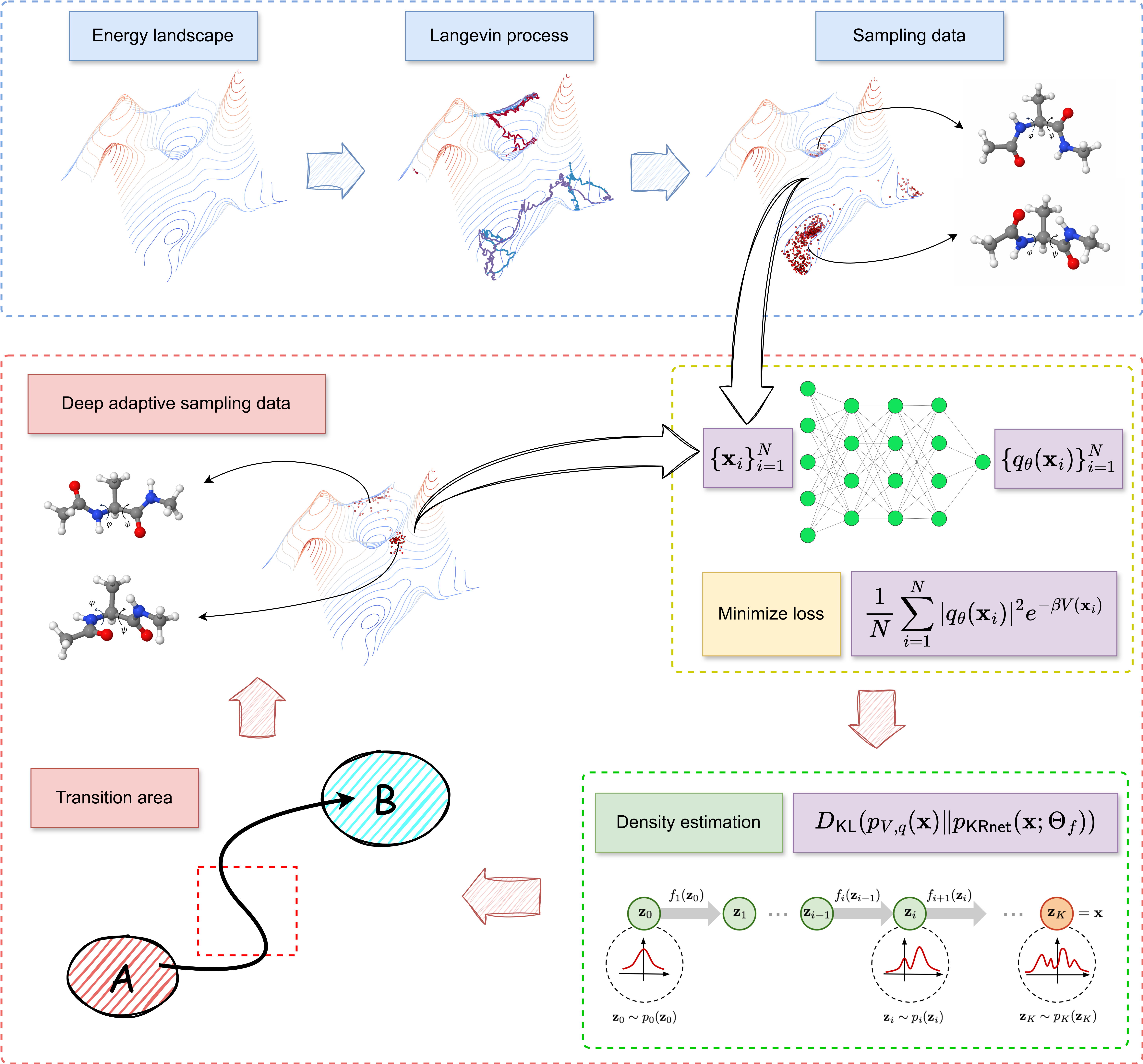}
	\end{center}
	\vspace{-13pt}
	\caption{\textbf{The schematic of DASTR for computing the committor function.} Training a deep neural network $q_{\mb{\theta}}$ to approximate the high-dimensional committor function must use a high-quality dataset (i.e. data points from the transition area). 
		Typically, the data points from Langevin dynamics are not in the transition state region since the transition between two metastable states is rare and difficult to sample. 
		The proposed DASTR method can adaptively generate effective data points on the transition area according to the information of the current approximate solution. 
		The key point is to define a sampling distribution $p_{V,q}$ dependent on the current approximate solution and the potential. Effective data points in the transition area are generated by sampling from $p_{V,q}$, which is achieved through training a deep generative model.}\label{fig:dastr_idea}
\end{figure}
Our adaptive sampling strategy is stated as follows. 
Let $\mathsf{S}_{0} = \{\mb{x}_{0,i} \}_{i=1}^{N_0}$  be a set of collocation points that are sampled from a given distribution $p_{0}(\mb{x})$ in $\Omega \backslash (A \cup B)$. Using $\mathsf{S}_{0}$, we minimize the empirical loss defined in \eqref{eq_vardiscrete} to obtain $q_{\mb{\theta}_1}$. With $q_{\mb{\theta}_1}$, we minimize the cross entropy in \eqref{eqn:ce_approx} to get $p_1 = p_{\mathsf{KRnet}}(\bx;\Theta_f^{*, (1)})$. A new set $\mathsf{S}^g_{1} = \{\mb{x}_{1,i} \}_{i=1}^{n_1}$ with $n_1 \leq N_0$ is generated by $f^{-1}_{\mathsf{KRnet}}(\mb{z}_i;\Theta_f^{*,(1)})$ (see \eqref{eqn:z_to_x}) to refine the training set. To be more precise, we replace $n_1$ points in $\mathsf{S}_0$ with $\mathsf{S}^g_1$ to get a new set $\mathsf{S}_1$. 
Then we continue to update the approximate solution $q_{\mb{\theta}_1}$ using 
$\mathsf{S}_{1}$ as the training set. 

\begin{algorithm}[!htb]
	\caption{DASTR}
	\label{alg_dastr}
	 \begin{algorithmic}[1]
		\Require Initial $q_{\mb{\theta}_0}$, maximum stage number $N_{\rm adaptive}$, maximum epoch number $N_e, N_e^{\prime}$, batch size $m, m^{\prime}$, initial training set $\mathsf{S}_{0} = \{\mb{x}_{0,i} \}_{i=1}^{N_0}$.
		\For{$k = 0:N_{\rm adaptive}-1$}
		\For {$i = 1:N_e$}
		\For{$l$ steps}
		\State Sample $m$ samples from $\mathsf{S}_{k}$.
		\State Update $q_{\mb{\theta}}(\mb{x})$ by descending the stochastic gradient of the discrete variational loss (see \eqref{eqn_update_theta}).
		\EndFor
		\EndFor
		\For {$i = 1:N_e^{\prime}$}
		\For {$l$ steps}
		\State Sample $m^{\prime}$ samples from the standard Gaussian distribution.
        \State Generate samples using \eqref{eqn:z_to_x}.
		\State Update $p_{\mathsf{KRnet}}(\bx;\Theta_f)$ by descending the stochastic gradient of $H(p_{V,q},p_{\mathsf{KRnet}})$ (see \eqref{eqn:ce_approx}).
		\EndFor
		\EndFor
		\State Refine the training set: use $p_{k+1}$ to get $\mathsf{S}_{k+1}$.
		\EndFor	
		\Ensure $q_{\mb{\theta}}$
	\end{algorithmic}
\end{algorithm}

In general, at the $k$-stage, suppose that we have $n_j$ samples $\mathsf{S}^g_{j} = \{\mb{x}_{j,i} \}_{i=1}^{n_j}$ from $p_j$ for $j = 1, \ldots, k$, where $p_j$ is the PDF model at the $j$-th stage and it can be trained by letting $p_{j-1} = p_{\mathsf{KRnet}}(\mb{x}_i;\Theta_f^{\prime})$ in \eqref{eqn:ce_approx}. The training set $\mathsf{S}_k$ at the $k$-th stage consists of $\mb{x}_{j,i}$. We use $\mathsf{S}_{k}$ to obtain $q_{\mb{\theta}_{k+1}}$ as
\begin{equation}\label{eqn_update_theta}
	\mb{\theta}_{k+1}= \arg \min_{\mb{\theta}} \sum\limits_{j=0}^k \frac{1}{n_j} \sum\limits_{i=1}^{n_j} \alpha_j \vert \nabla q_{\mb{\theta}}(\mb{x}_{j,i}) \vert^2 \frac{e^{-\beta V(\mb{x}_{j,i})}}{p_{j}(\mb{x}_{j,i})},
\end{equation}
where $q_{\mb{\theta}}$ is initialized as $q_{\mb{\theta}_{k}}$, $\alpha_j = n_j/\sum_{j=0}^{k}{n_j}$ is a weight to balance the different distributions $p_j$, and $n_0$ is the number of points kept in $\mathsf{S}_0$ at the $k$-th stage.
Starting with $p_{k} = p_{\mathsf{KRnet}}(\bx;\Theta_f^{*,(k)})$, the density model $p_{\mathsf{KRnet}}(\bx;\Theta_f)$ is updated by \eqref{eqn:ce_approx} to get $p_{k+1}$.
A new set $\mathsf{S}^g_{k+1} = \{\mb{x}_{k+1, i} \}_{i=1}^{n_{k+1}}$ of collocation points is generated by \eqref{eqn:z_to_x}. We then use $\mathsf{S}^g_{k+1}$ to refine the training set to get $\mathsf{S}_{k+1}$. We repeat the above procedure to obtain Algorithm \ref{alg_dastr} for the deep adaptive sampling on transition paths. We call this method DASTR for short. The main idea of our algorithm is also illustrated in Figure \ref{fig:dastr_idea}.

\subsection{DASTR in the Latent Space}
\begin{figure}[!htb]
	\centering
	\subfloat[][Molecules generated using coordinates of heavy atoms as the input to KRnet. \label{fig:unreasonable_configs}]{\includegraphics[width=.40\textwidth]{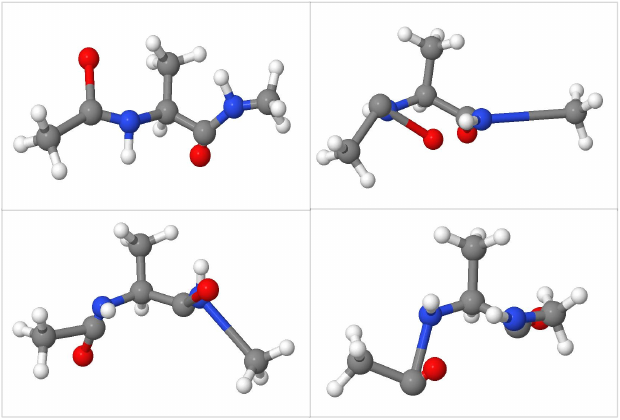}}	\hspace{2em}
    \subfloat[][Molecules generated using latent CVs as the input to KRnet. \label{fig:reasonable_configs}]{\includegraphics[width=.40\textwidth]{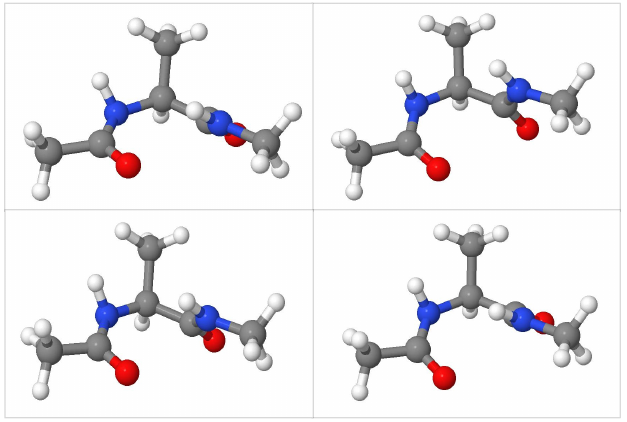}}
	\caption{Molecular configurations generated by two different settings in DASTR: (a) the inputs of KRnet are the coordinates of heavy atoms (b) the inputs of KRnet are the latent CVs. The hydrogen atoms are completed by the software package PyMOL \citep{schrodinger2015pymol}. This figure demonstrates that using the latent collective variables to conduct DASTR is more effective.}
    \label{fig:re_unreasonable_configs}
\end{figure}

For complex systems, such as protein molecules, directly applying DASTR will result in the generation of physically unreasonable molecular configurations during the adaptive sampling procedure. The reason behind this is the strong correlation among the atomic coordinates required by physically reasonable protein structures. As a result, directly using the atomic coordinates as input to the KRnet may fail to capture the interatomic relationships effectively. 
This observation is demonstrated in Figure~\ref{fig:re_unreasonable_configs}. The molecular configurations in the left plot, which are almost physically unreasonable, are sampled from a trained KRnet in the original high-dimensional space, while the molecular configurations in the right plot, which are physically consistent, are sampled using latent collective variables as discussed later in section \ref{sec_lcvs_ae}. 

To resolve this issue, we resort to sampling in the latent space, where we consider two strategies: one is based on the collective variables (CVs) method \cite{fiorin2013using} (see section \ref{sec_hpcvs_us}), and the other is based on autoencoders (see section \ref{sec_lcvs_ae}). CVs refer to variables that can capture critical information about molecular structures. For example, the dihedral angles of the backbone atoms or distance between atoms can be selected as the CVs in protein systems. CVs can help reduce the computational complexity and enhance the sampling correctness. Moreover, we propose to use an autoencoder to automatically select latent CVs, which in general do not have explicit physical meanings, and generate physically reasonable molecular configurations using these latent CVs.

The basic idea of the collective variables method is to replace the original coordinates with some collective variables $\mb{s}(\mb{x}) = [s_1(\mb{x}), \ldots, s_m(\mb{x})]^\top$ with $m \ll d$, where $d$ is the dimensionality of $\mb{x}$. Then we can restrict our attention to the collective variables during the adaptive sampling procedure: 
\begin{equation}\label{eq_pvq_cv}
p_{V,q}(\mb{s}(\mb{x})) = p_{V, q}(\mb{x}),
\end{equation}
where the $p_{V, q}(\mb{x})$ corresponds to the term defined in equations \eqref{eq_committor_pdf1} and \eqref{eq_committor_pdf2}. Since the collective variables can capture the essential structural features of molecules, one can take adaptively sampling step on the collective variables $\mb{s}(\mb{x})$ as illustrated in Algorithm \ref{alg_dastr}. To generate samples in the latent space, we need to train KRnet using the CVs as input to learn the probability distribution in the latent space. Similar to the discussions in section \ref{sec_dastr_sample}, training KRnet can be performed by minimizing the cross entropy loss defined in the latent space. This way, the deep generative model is used to generate samples of the collective variables instead of the coordinates $\mb{x}$. After generating the collective variables, one can do some post-processing steps to obtain new samples of $\mb{x}$. This will reduce the probability of generating nonphysical samples.

If there is no prior information for selecting the proper collective variables, we use an autoencoder to learn some latent variables from the data and use them as the collective variables. The overall procedure along this line is summarized in Algorithm \ref{alg_dastr_cv}. 

\begin{algorithm}[!htb]
	\caption{DASTR in the latent space}
	\label{alg_dastr_cv}
	 \begin{algorithmic}[1]
		\Require Initial $q_{\mb{\theta}_0}$, maximum stage number $N_{\rm adaptive}$, maximum epoch number $N_e, N_e^{\prime}$, batch size $m, m^{\prime}$, initial training set $\mathsf{S}_{0} = \{\mb{x}_{0,i} \}_{i=1}^{N_0}$.
        \If{Using autoencoder}
        \State Train the autoencoder using $\mathsf{S}_0$.
        \EndIf
		\For{$k = 0:N_{\rm adaptive}-1$}
		\For {$i = 1:N_e$}
		\For{$l$ steps}
		\State Sample $m$ samples from $\mathsf{S}_{k}$.
		\State Update $q_{\mb{\theta}}(\mb{x})$ by descending the stochastic gradient of the discrete variational loss (see \eqref{eqn_update_theta}).
		\EndFor
		\EndFor
		\For {$i = 1:N_e^{\prime}$}
		\For {$l$ steps}
		\State Sample $m^{\prime}$ samples from the standard Gaussian distribution.
        \If{Using autoencoder}
		\State Update $p_{\mathsf{KRnet}}(\mb{s}(\bx);\Theta_f)$ by descending the stochastic gradient of $H(p_{V,q},p_{\mathsf{KRnet}})$ using  \eqref{eqn:ce_approx_cv_auto}.
        \Else
        \State Update $p_{\mathsf{KRnet}}(\mb{s}(\bx);\Theta_f)$ by descending the stochastic gradient of $H(p_{V,q},p_{\mathsf{KRnet}})$ using \eqref{eqn:ce_approx_cv_us})
        \EndIf
		\EndFor
		\EndFor
        \State Generate new samples of the latent collective variables by the trained KRnet.
        \State Use the pretrained decoder to get new samples of $\mb{x}$. 
		\State Refine the training set to get $\mathsf{S}_{k+1}$.
		\EndFor	
		\Ensure $q_{\mb{\theta}}$
	\end{algorithmic}
\end{algorithm}

\subsubsection{Hand-picking CVs with Umbrella Sampling}\label{sec_hpcvs_us}
We first consider that the explicit collective variables are available. In this scenario, the dihedral angles of the backbone atoms are selected as CVs \citep{li2019computing}. As discussed above, we need to ensure that the samples obey the molecular configurations during the adaptive sampling procedure.

It is straightforward to train a KRnet to model the distribution in terms of collective variables $\mb{s}$. The KRnet that maps the collective variables $\mb{s}$ to a standard Gaussian is obtained by minimizing the following cross entropy
\begin{equation}\label{eqn:ce_approx_cv_us}
    H(p_{V, q}, p_{\mathsf{KRnet}}) \approx -\frac{1}{N}\sum_{i=1}^{N} \frac{p_{V, q}(\mb{s}(\mb{x}_i))}{p_{\mathsf{IS}}(\mb{s}(\mb{x}_i))}\log p_{\mathsf{KRnet}}(\mb{s}(\mb{x}_i);\Theta_f),
\end{equation}
where $p_{\mathsf{IS}}(\mb{s}(\mb{x}_i)) = e^{-\beta V_{\text{modified}}(\mb{x}_i)}$ and each $\mb{x}_i$ is a sample drawn from the previous step. The generation of new samples for $\mb{x}$ is achieved in two steps: we first generate samples for the collective variables $\mb{s}$ using the trained KRnet, and then sample $\mb{x}$ that satisfies $\mb{s}(\mb{x})\approx \mb{s}$ using umbrella sampling~\citep{kastner2011umbrella} (see \ref{sec_appendix_us} for more details). 

The potential function $V_{\text{modified}}$ is used to simulate the SDE to generate new samples 
\begin{equation*}
	V_{\text{modified}}(\mb{x}) = V(\mb{x}) + V_{\text{US}}(\mb{x}),
\end{equation*}
where $V$ is the original potential in \eqref{eq_committor_sde} and $V_{\text{US}}(\mb{x})$ is the the umbrella potential with the following form
\begin{equation}\label{Umbrella_potential}
  V_{\text{US}}(\mb{x}) = \frac{1}{2} \sum_{i=1}^{m} k_{\text{us}} (s_{i}(\mb{x}) - s_{i}(\mb{x}_0))^2. 
\end{equation}
Here, $s_{i}(\mb{x}_0)$ is the target CVs generated by the trained KRnet, $s_{i}(\mb{x})$ represents the CVs with respect to $\mb{x}$, $m$ is the number of CVs, and $k_{\text{us}}$ is the force constant. 
We perform a rapid iterative process of umbrella sampling to transfer the CVs to the target region, and finally sample near the target CVs in the modified potential. This ensures the physical validity of the molecular configurations during the adaptive sampling procedure. However, selecting proper collective variables requires additional domain-specific knowledge, which is not a trivial task. Additionally, this strategy for implementing adaptive sampling in the latent space still requires simulating the SDE, which limits its sampling efficiency.

\subsubsection{Latent CVs with Autoencoder}\label{sec_lcvs_ae}
In this part, we provide an alternative way of using an autoencoder to automatically select the latent variables as the collective variables. The autoencoder can be trained before the first stage in Algorithm \ref{alg_dastr_cv} using the data from metadynamics. After training, the autoencoder is fixed during the adaptive sampling procedure. 

The configurations of molecular systems are primarily determined by the positions of the heavy atoms and the positions of the hydrogen atoms can be inferred from the positions of the heavy atoms. Based on this observation, we selected the coordinates of all the heavy atoms of molecules from $\mathsf{S}_0$ as the dataset for training the autoencoder. The autoencoder consists of two parts: an $\mathsf{encoder} = \mb{s}(\mb{x})$ and a $\mathsf{decoder} = \mb{S}(\mb{s}(\mb{x}))$. Both the encoder and decoder are modeled by neural networks. Training the autoencoder aims to minimize the mean squared error 
\begin{equation*}
\frac{1}{N} \sum\limits_{i=1}^N (\mb{S}(\mb{s}(\mb{x}_i)) - \mb{x}_i)^2.
\end{equation*}
Once the autoencoder is trained, the latent CVs can be obtained by the encoder. 
To this end, we utilize KRnet to learn the distribution of the latent CVs by minimizing the following cross entropy with respect to the latent CVs
\begin{equation}\label{eqn:ce_approx_cv_auto}
    H(p_{V, q}, p_{\mathsf{KRnet}}) \approx -\frac{1}{N}\sum_{i=1}^{N} \frac{p_{V, q}(\mb{s}(\mb{x}_i))}{p_{\mathsf{KRnet}}(\mb{s}(\mb{x}_i);\Theta_f^{\prime})}\log p_{\mathsf{KRnet}}(\mb{s}(\mb{x}_i);\Theta_f),
\end{equation}
where the parameters $\Theta_f^{\prime}$ can be chosen from the previous step.

\begin{figure}[!htb]
	\begin{center}
		\includegraphics[width=.96\textwidth]{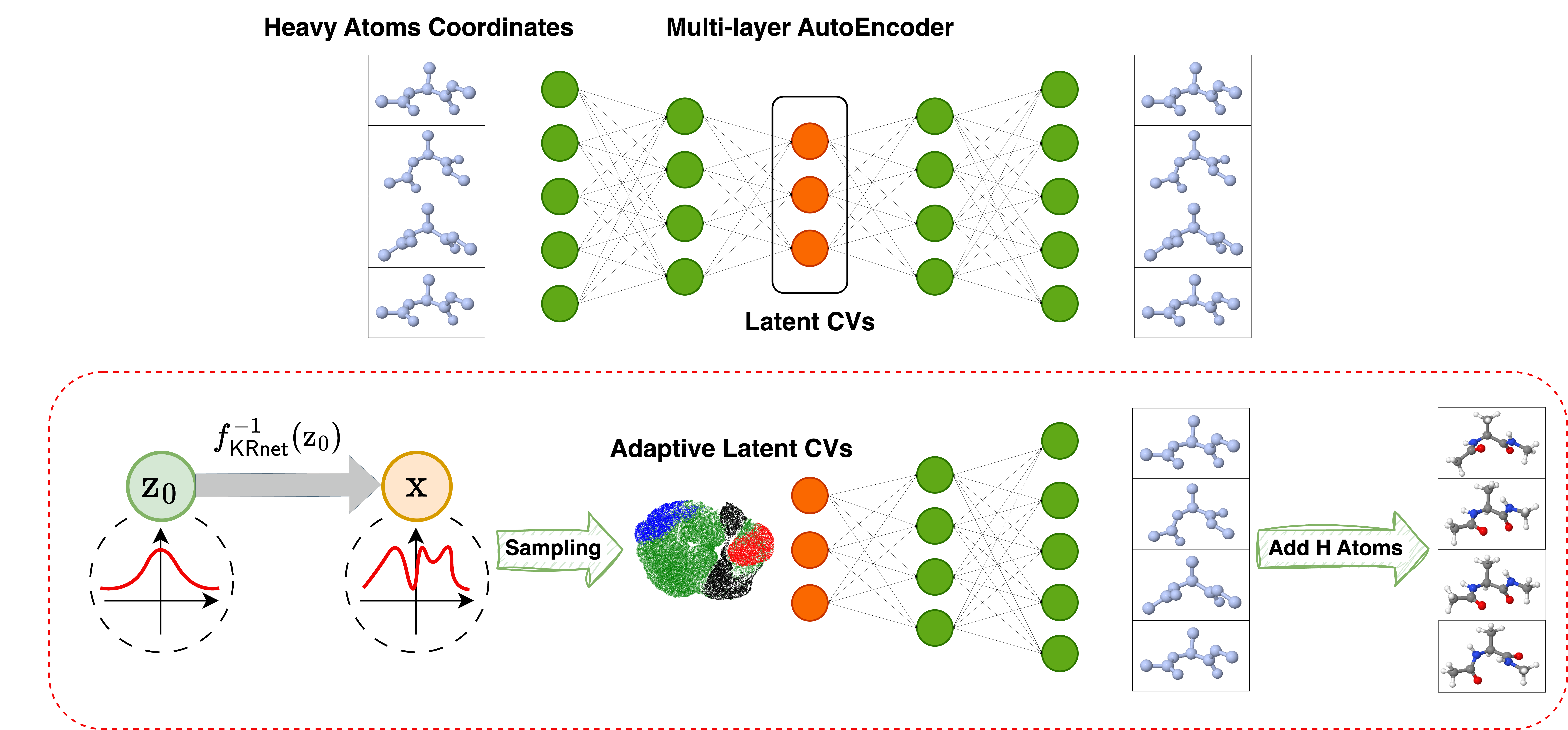}
	\end{center}
	\caption{\textbf{The schematic of adaptive sampling in the latent space.} We first train an autoencoder to obtain the latent variables as the collective variables (CVs), and then use KRnet to approximate the distribution of the CVs. After training KRnet, we use a random sample $\mb{z}_0$ from the standard Gaussian distribution to generate a new sample of latent CVs. We can feed this new sample of latent CVs into the decoder to obtain a new sample of molecules after the post-processing step. Such a new sample of molecules is located in the transition state region. The autoencoder not only provides an effective way to automatically choose the collective variables, but also enhances the sampling efficiency of molecules in the transition state region.}\label{fig:autoencder_flow}
\end{figure}

Once we have the trained KRnet in hand, we can generate samples $\mb{s}$ in the latent space. These samples are subsequently decoded using the pretrained decoder to reconstruct the positions of all the heavy atoms. The hydrogen atoms are automatically completed using the software package PyMOL \citep{schrodinger2015pymol}. Finally, we calculate the potential energy of the generated molecular configurations to exclude those samples with excessively high potential energies, thus avoiding the generation of physically unreasonable configurations. The generated molecular configurations are illustrated in Figure~\ref{fig:reasonable_configs}. The proportion of reasonable configurations generated by this method exceeds $97\%$ (details can be found in section \ref{sec_numexp_ad_latent_cv}). The computation process is illustrated in Figure \ref{fig:autoencder_flow}.

\begin{remark}
	The key point here is that the autoencoder helps us automatically obtain the latent collective variables that reflect the molecular configuration, which serve as the input of KRnet, without the need of hand-picking physical CVs. In section \ref{sec_numexp_ad_explicit_CVs}, we use KRnet to learn the distribution corresponding to the physical CVs and employ umbrella sampling to generate samples of molecules based on these physical CVs. However, this process consumes significant time and computational resources because umbrella sampling is still based on the SDE simulation. In contrast, the autoencoder explores latent CVs, allowing us to break free from the reliance on physical CVs and the associated SDE-based sampling methods. Moreover, the decoder can quickly reconstruct the molecular structure, significantly improving the computational efficiency. We compare the sampling time of the two methods in section \ref{sec_numexp_ad_latent_cv}.
\end{remark}

\section{Numerical Study}\label{sec_numexp}
We conduct three numerical experiments to demonstrate the effectiveness of the proposed method. The first one is a 10-dimensional rugged Mueller potential problem, the second one is a 20-dimensional standard Brownian motion problem, and the last one is the alanine dipeptide problem with the dimension $d = 66$. The performance of DASTR with the collective variables method and the autoencoder method is investigated using the alanine dipeptide problem. The detailed settings of numerical experiments are provided in \ref{sec_appdix_numdetail}.

\subsection{Rugged Mueller Potential}
We consider the extended rugged Mueller potential embedded in the 10-dimensional space, which is a well-known test problem in computational chemical physics \citep{li2019computing, li2022semigroup}. The extended rugged Mueller potential is given by $V(\mb{x}) = V_{\mathrm{rm}}(x_1, x_2) + 1/(2 \sigma^2) \sum_{i=3}^{10} x_i^2$,
where $\mb{x} \in\mathbb{R}^{10}$ and $V_{\mathrm{rm}}(x_1, x_2)$ is the rugged Mueller potential defined in $[-1.5, 1] \times [-0.5, 2]$
\begin{equation*}
	V_{\mathrm{rm}}(x_1, x_2) = \sum\limits_{i=1}^4 D_i e^{a_i(x_1 - \xi_i)^2 + b_i(x_1 - \xi_i)(x_2 - \eta_i) + c_i(x_2 - \eta_i)^2} + \gamma  \mathrm{sin}(2k\pi x_1) \mathrm{sin}(2k \pi x_2).
\end{equation*}
We set $\sigma = 0.05$ as in \citep{li2019computing}, and the other parameters are set to be the same as in \citep{lai2018point}. The inverse temperature is set to $\beta = 1/10$. In this test problem, the two metastable sets $A$ and $B$ are two cylinders with centers $[x_1, x_2] = [-0.558, 1.441]$ and $[x_1, x_2] = [0.623, 0.028]$ respectively with radius $0.1$. In this setting, the solution of this 10-dimensional problem is the same as that of the two-dimensional rugged Mueller potential, i.e., $q(\mb{x}) = q_{\mathrm{rm}}(\mb{x})$ \citep{li2019computing, li2022semigroup}. So, we can use the finite element method implemented in FEniCS \citep{alnaes2015fenics,logg2012automated} to obtain a reference solution to evaluate the performance. For comparison, we also implement the artificial temperature method \citep{li2019computing} as the baseline model. Here we define the $L^{2}$ relative error $\norm{ \mb{q}_{\mb{\theta}} - \mb{q}_{\rm{ref}}}{2}/\norm{\mb{q}_{\rm{ref}}}{2}$,
where $\mb{q}_{\mb{\theta}}$ and $\mb{q}$ denote two vectors whose elements are the function values of $q_{\mb{\theta}}$ and $q_{\rm{ref}}$ at some grids respectively. The settings of neural networks and training details can be found in \ref{sec_appendix_setting_rmp}.

\begin{figure}[!htb]
	\centering
	\subfloat[][Samples from SDE.
	\label{fig:rm_10d_points_sde}]{\includegraphics[width=.27\textwidth]{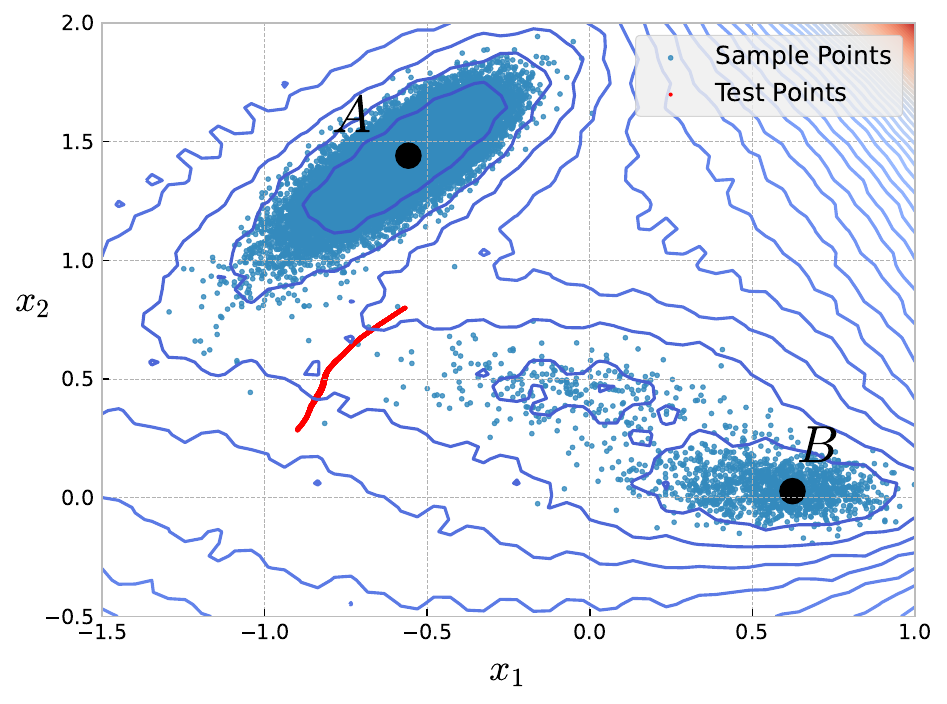}}\quad
	\subfloat[][Samples from SDE using the artificial temperature method, $\beta = 1/20$. \label{fig:rm_10d_points_sdeT20}]{\includegraphics[width=.27\textwidth]{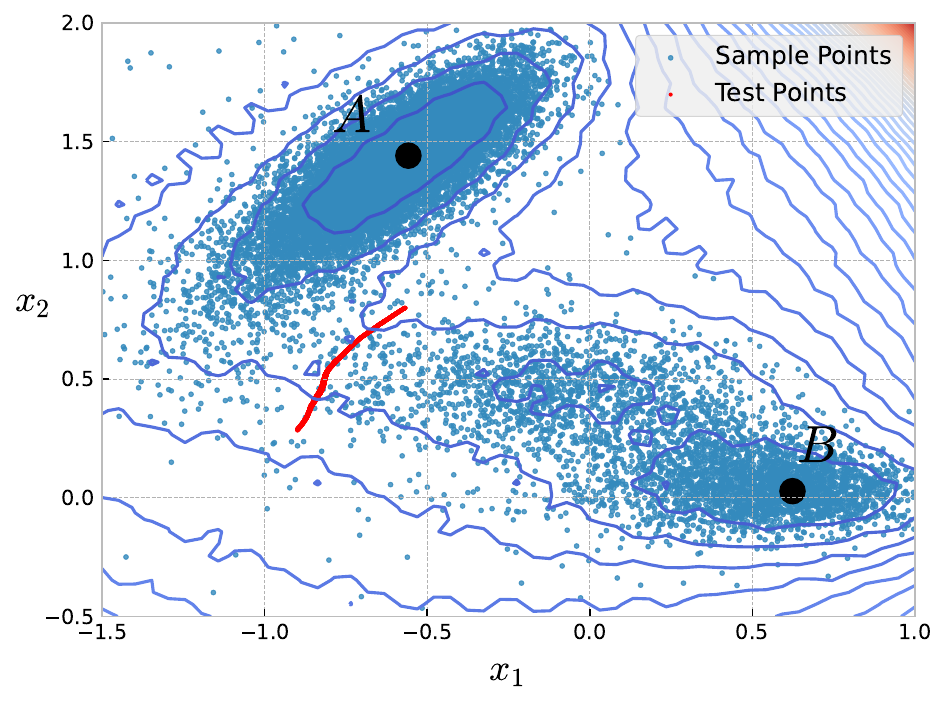}}\quad
	\subfloat[][DASTR, samples at the $1$-th stage. \label{fig:rm_10d_points_das_2}]{\includegraphics[width=.27\textwidth]{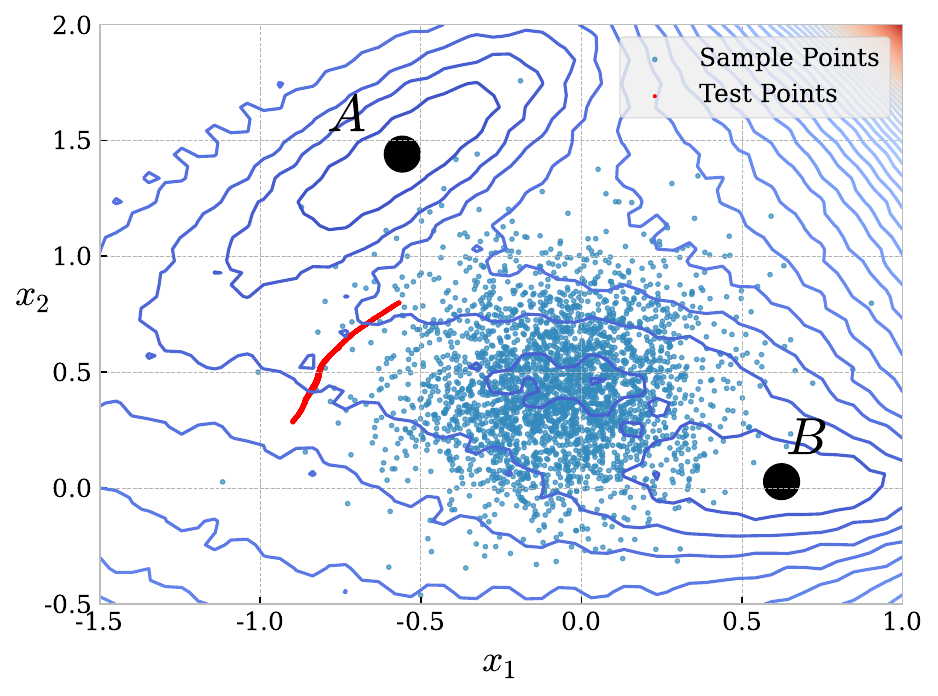}}\\
	
	\subfloat[][DASTR, samples at the $5$-th stage.
	\label{fig:rm_10d_points_das_5}]{\includegraphics[width=.27\textwidth]{rm_10d_points_das_5.pdf}}\quad
	\subfloat[][DASTR, samples at the $15$-th stage.\label{fig:rm_10d_points_das_15}]{\includegraphics[width=.27\textwidth]{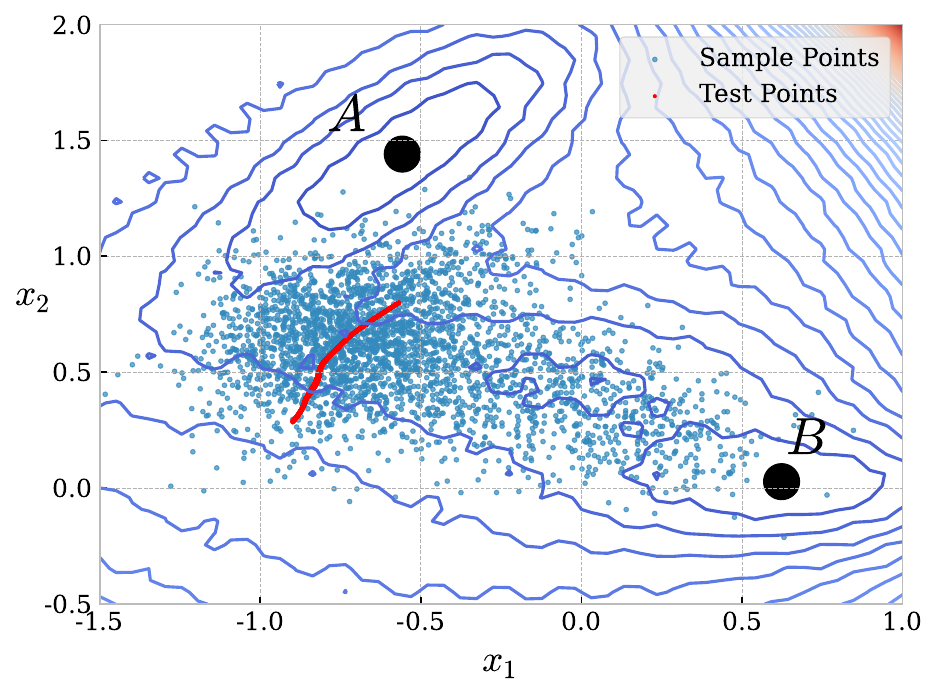}}\quad
	\subfloat[][DASTR, samples at the $29$-th stage. \label{fig:rm_10d_points_das_30}]
	{\includegraphics[width=.27\textwidth]{rm_10d_points_das_30.pdf}}
	\caption{DASTR, samples for the 10-dimensional rugged Mueller potential problem. The red line denotes the test points from the $1/2$-isosurface ($q \approx 1/2$) projected onto the $x_1$-$x_2$ plane. }
	\label{fig:rm_10d_sample}
\end{figure}

Figure~\ref{fig:rm_10d_sample} shows the samples from different sampling strategies, where these samples are projected onto the $x_1$-$x_2$ plane. Specifically, Figure~\ref{fig:rm_10d_points_sde} shows the samples generated by SDE defined in \eqref{eq_committor_sde}. It can be seen that the samples from SDE are located around the two metastable states $A$ and $B$, which are ineffective for  approximating the committor function. Figure~\ref{fig:rm_10d_points_sdeT20} shows the samples from SDE with the artificial temperature method. While more samples show up in the transition state region compared with Figure~\ref{fig:rm_10d_points_sde}, 
there still does not exist sufficient information in the dataset to capture the committor function well. Our method is able to provide effective samples in the transition area. As shown in Figures~\ref{fig:rm_10d_points_das_2}-\ref{fig:rm_10d_points_das_30}, the evolution of the training set with respect to adaptivity iterations $k = 1, 5, 15, 29$ is presented, where we randomly select $5000$ samples in the training set for visualization. Obviously, such samples are distributed in the transition state region ($\Omega \backslash (A \cup B)$), which is desired for approximating the committor function.
\begin{figure}[!htb]
	\centering
	\subfloat[][Error behavior \label{fig:10drm_error_epoch}]{\includegraphics[width=.225\textwidth,height=.161\textwidth]{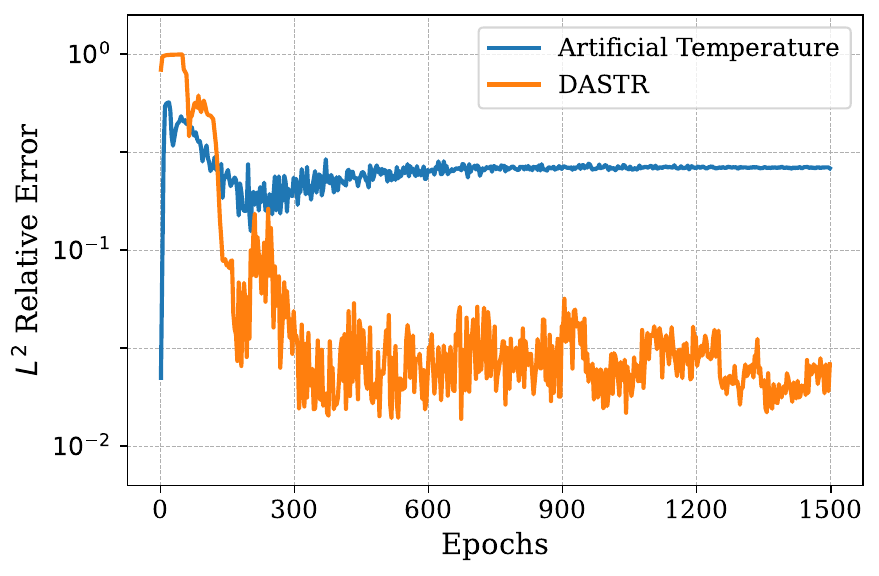}}\quad
	\subfloat[][Reference \label{fig:sol_ref}]{\includegraphics[width=.215\textwidth]{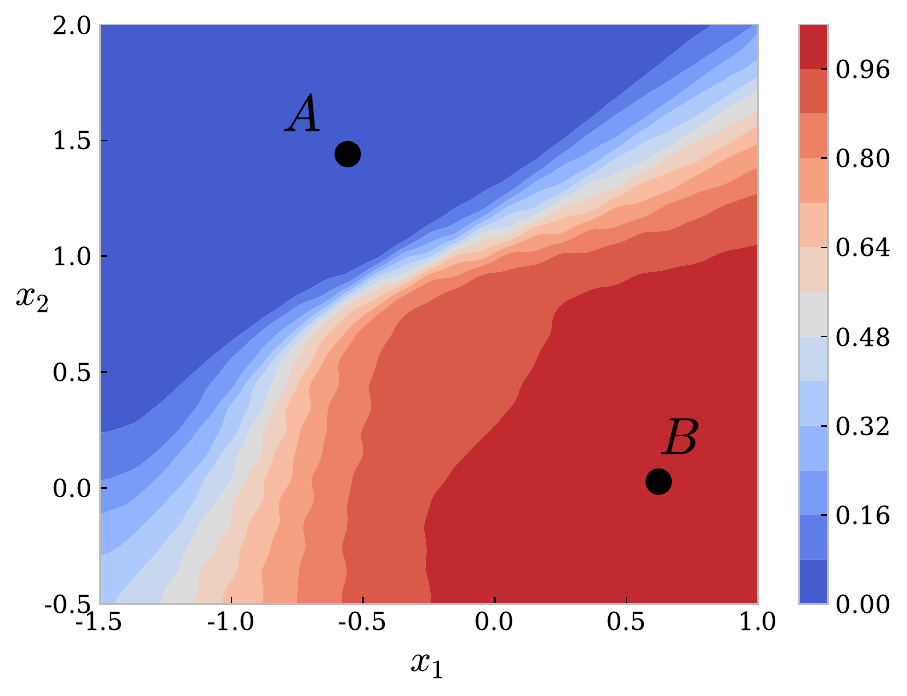}}\quad
	\subfloat[][DASTR \label{fig:sol_dastr}]{\includegraphics[width=.215\textwidth]{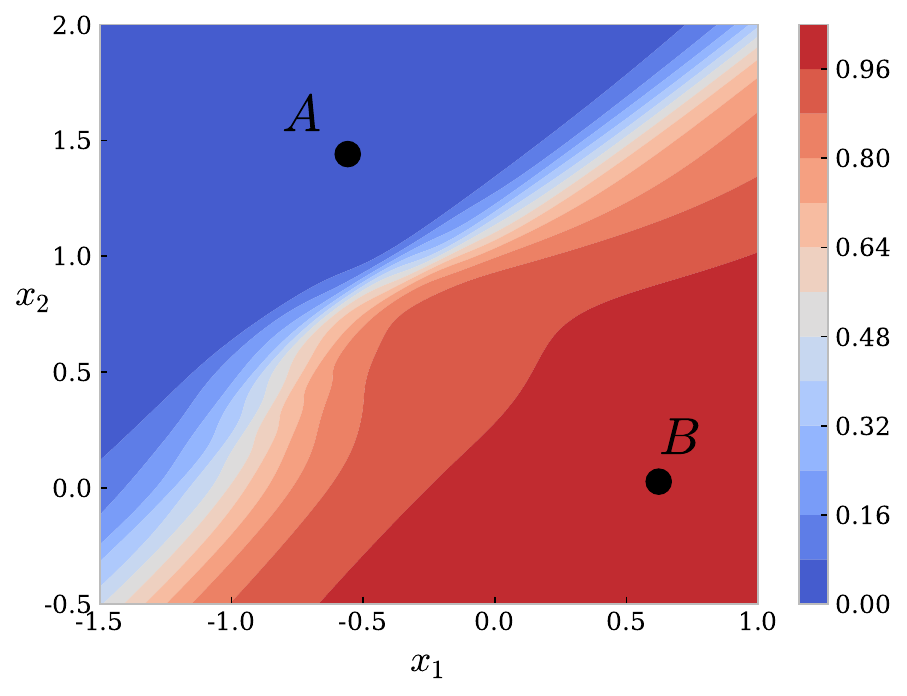}}\quad
	\subfloat[][The artificial temperature method \label{fig:sol_sde}]
	{\includegraphics[width=.215\textwidth]{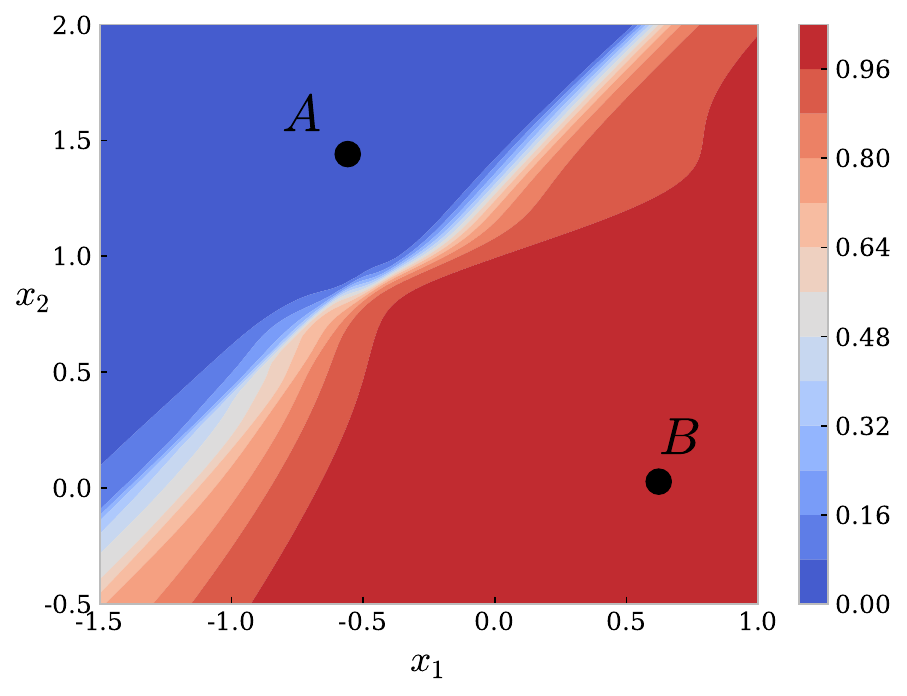}}\quad
	\caption{Solutions, $10$-dimensional rugged Mueller potential test problem. }
	\label{fig:rm_10d_sol}
\end{figure}

\begin{figure}[!htb]
	\begin{center}
		\includegraphics[width=.46\textwidth]{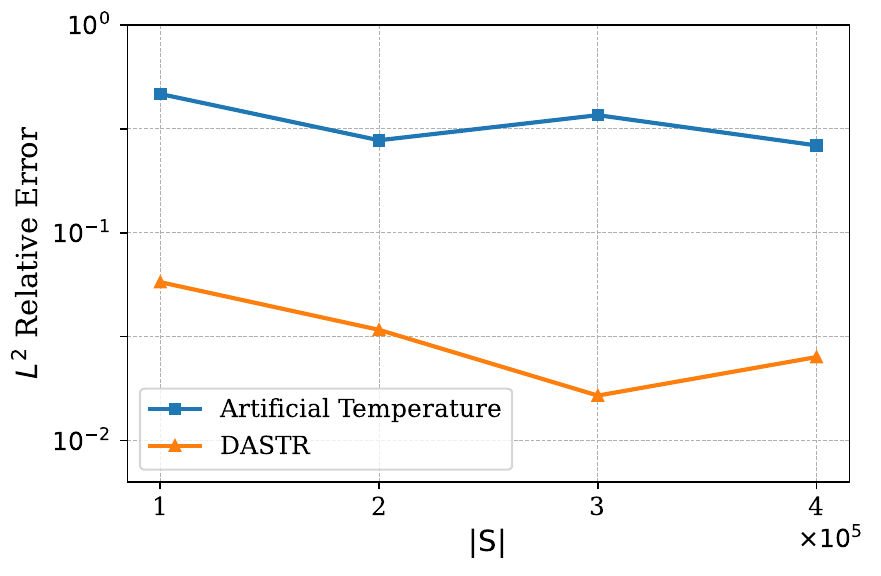}
	\end{center}
	\vspace{-13pt}
	\caption{The error w.r.t. sample size $\vert \mathsf{S} \vert$.}
	\label{fig:rm_10d_error}
\end{figure}

Figure~\ref{fig:10drm_error_epoch} shows the error behavior of different methods. In Figure \ref{fig:sol_ref}-\ref{fig:sol_sde}, we compare the reference solution  $q_{\rm{ref}}$ obtained by the finite element method, the DASTR solution given by $4 \times 10^5$ samples and the approximate solution given by $4 \times 10^5$ samples with the artificial temperature method. Figure~\ref{fig:rm_10d_error} shows the relative errors with respect to different sample sizes. From Figure~\ref{fig:rm_10d_error}, it is seen that the DASTR method is much more accurate than the method of sampling from dynamics. Due to the difficulty of sampling in the transition state region using SDE with the artificial temperature method, the solution obtained through the artificial temperature method fails to accurately capture the information of the committor function in the transition state region. To further investigate the performance of the proposed method,  in Table~\ref{tab:rm_10d_error}, we show the $L^{2}$ relative errors of neural networks with varying numbers of neurons subject to different sample sizes. Here, we sample $12099$ points near the $1/2$-isosurface ( $q(\mb{x}) \approx 0.5$ ) to compute the relative error. Our DASTR method is one order of magnitude more accurate than the baseline method in all settings. 

\begin{table}[!htb]
	\scriptsize
	\centering
	\caption{$10$-dimensional rugged Mueller potential test problem: errors for different settings of neural networks and sampling strategies. We take $4$ independent runs to compute the error statistics (mean $\pm$ standard deviation).}
	\renewcommand{\arraystretch}{1.5}  %
	\begin{tabular}{ccccc}
		\toprule
		& & \multicolumn{3}{c}{\textbf{Number of Neurons in Hidden Layer}} \\  
		\hline
		\textbf{Sampling Method} & $\vert \mathsf{S} \vert$ & \textbf{20} & \textbf{50} & \textbf{100} \\ 
		\hline
		\multirow{4}{*}{\makecell[c]{SDE with the \\ artificial temperature method}} & $1 \times 10^5$ & 0.5446 $\pm$ 0.0724 &  0.4693 $\pm$ 0.0627 & 0.4023 $\pm$ 0.0819  \\
		& $2 \times 10^5$ & 0.3183 $\pm$ 0.0592 &  0.2677 $\pm$ 0.0708 & 0.3063 $\pm$ 0.0477  \\
		& $3 \times 10^5$ & 0.2717 $\pm$ 0.0487 &  0.2780 $\pm$ 0.0584  & 0.3955 $\pm$ 0.0311 \\
		& $4 \times 10^5$ & 0.3822 $\pm$ 0.0555 & 0.3019 $\pm$ 0.0649 & 0.3822 $\pm$ 0.1213  \\
		\hline
		\multirow{4}{*}{\centering DASTR (this work)} & $1 \times 10^5$ &0.0620 $\pm$ 0.0070 & 0.0602 $\pm$ 0.0113 & 0.0615 $\pm$ 0.0071  \\
		&  $2 \times 10^5$ &0.0498 $\pm$ 0.0102 & 0.0443 $\pm$ 0.0049 & 0.0310 $\pm$ 0.0024  \\
		&  $3 \times 10^5$ &0.0386 $\pm$ 0.0089 & 0.0412 $\pm$ 0.0091& 0.0172 $\pm$ 0.0028   \\
		&  $4 \times 10^5$ &0.0371 $\pm$ 0.0056 & 0.0343 $\pm$ 0.0065  & 0.0206 $\pm$ 0.0052 \\
		\bottomrule
	\end{tabular}
	\label{tab:rm_10d_error}
\end{table}

\subsection{Standard Brownian Motion}
In this test problem, we consider the committor function under the standard Brownian motion \citep{hartmann2019variational, jml23nikolas}.
For a stochastic process $(\mb{X}_t)_{t \geq 0} \in \mathbb{R}^d$, which is a standard Brownian motion starting at \(\mb{x} \in \mathbb{R}^d\), that is, $\mb{X}_t = \mb{x} + \mb{W}_t$, corresponding to \(\nabla{V(\mb{X}_t)} = 0\) and \(\beta = 1/2\) in \eqref{eq_committor_sde}.
The two metastable sets $A$ and $B$ are defined as $A = \{ \mb{x} \in \mathbb{R}^d : \norm{\mb{x}}{2} < a \}, B = \{ \mb{x} \in \mathbb{R}^d : \norm{\mb{x}}{2} > b \}$
with \(b > a > 0\). With these settings, for $d \geq 3$, there exists an analytical solution $q(\mb{x}) = (a^2 - \norm{\mb{x}}{2}^{2-d} a^2)/(a^2 - b^{2-d} a^2)$. 
In this test problem, we set $d = 20$ and $a = 1, b = 2$. The settings of neural networks and training details can be found in \ref{sec_appendix_setting_sbm}. Since the solution to this test problem cannot be projected onto the low-dimensional space, we here compare different sampling methods by computing the $L^{2}$ relative error at a validation set with $5000$ data points along a curve $\{(\kappa, \ldots, \kappa)^\top : \kappa \in [a/\sqrt{d}, b/\sqrt{d}]\}$ \citep{jml23nikolas}.

\begin{figure}[!htb]
	\centering
	\subfloat[][Solutions, $\vert \mathsf{S} \vert = 2 \times 10^4$. \label{fig:sol_20d_sbm}]{\includegraphics[width=.27\textwidth]{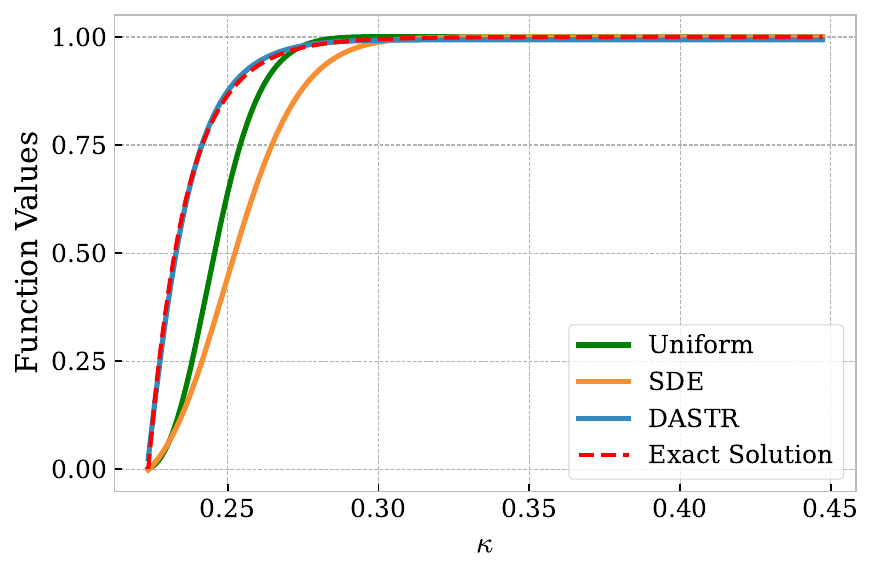}}\quad
	\subfloat[][The error evolution, $\vert \mathsf{S} \vert = 2\times 10^{4}$. \label{fig:error_epoch_20d_sbm}]{\includegraphics[width=.27\textwidth]{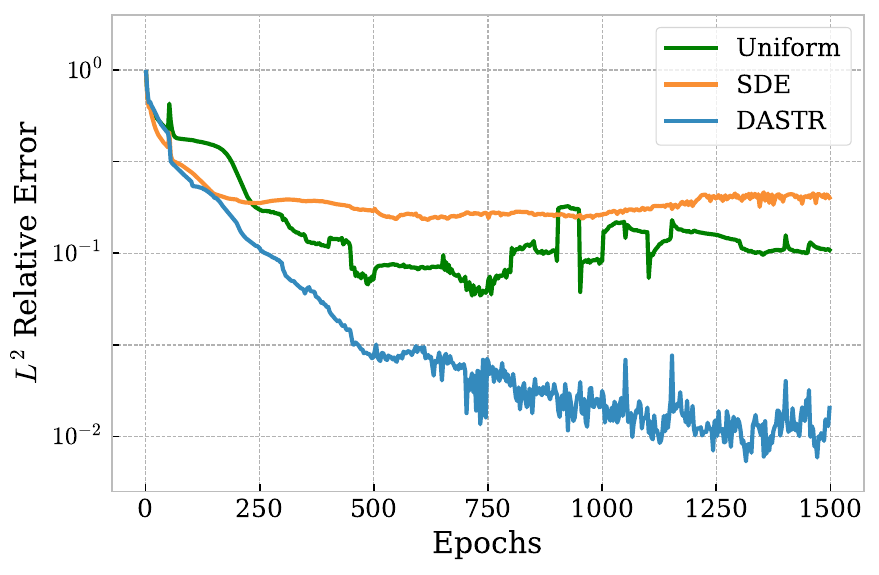}}\quad
	\subfloat[][The error w.r.t. sample size. \label{fig:error_sz_20d_sbm}]{\includegraphics[width=.27\textwidth]{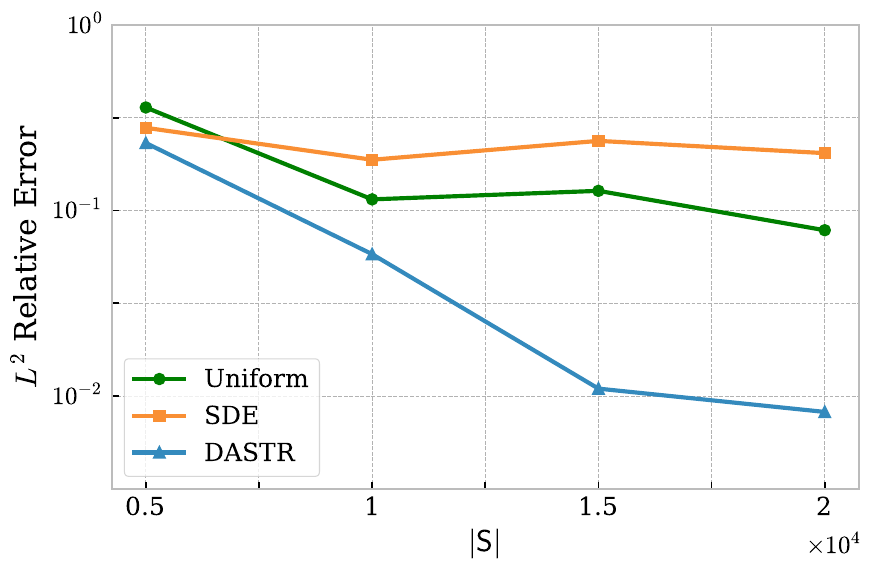}}
	\caption{Solutions evaluated along a curve and the behavior of relative errors, $20$-dimensional  standard Brownian motion test problem. The relative error is computed at the points along the curve $\{(\kappa, \ldots, \kappa)^\top : \kappa \in [a/\sqrt{d}, b/\sqrt{d}]\}$. }
	\label{fig:brownian_sol}
\end{figure}

Figure~\ref{fig:brownian_sol} shows the results of the $20$-dimensional standard Brownian motion test problem. Specifically, Figure~\ref{fig:sol_20d_sbm} shows the solutions obtained by different sampling methods, where it can be seen that the DASTR solution is more accurate than those of other sampling strategies. Figure~\ref{fig:error_epoch_20d_sbm} shows the behavior of relative errors during training, where DASTR performs better than the uniform sampling strategy and SDE. Figure~\ref{fig:error_sz_20d_sbm} shows the relative errors for the uniform sampling method, SDE, and DASTR, where different numbers of samples are tested. From Figure~\ref{fig:error_sz_20d_sbm}, it is clear that, as the number of samples increases, the relative error of DASTR decreases more quickly than those of SDE and the uniform sampling strategy.

\begin{figure}[!ht]
	\centering
	\subfloat[][Uniform samples. \label{fig:brow_uniform_points}]{\includegraphics[width=.28\textwidth]{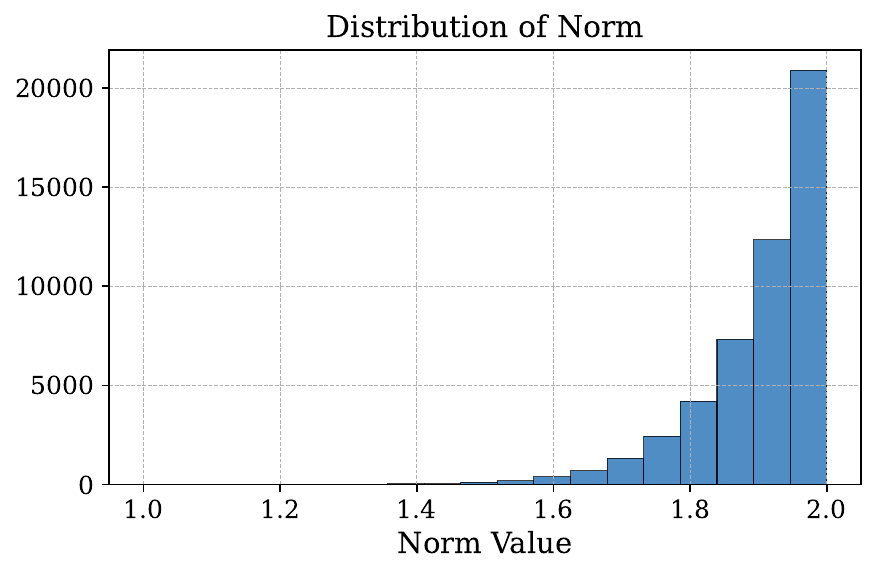}}\quad
	\subfloat[][Samples from SDE. \label{fig:brow_sde_points}]{\includegraphics[width=.28\textwidth]{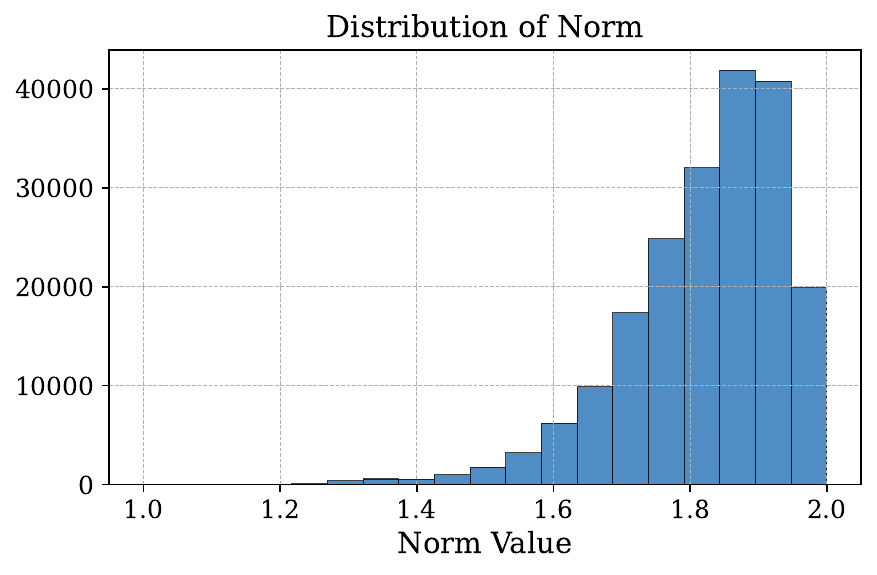}}\quad
	\subfloat[][DASTR, $k = 2$. 
	\label{fig:brow_das2_points}]{\includegraphics[width=.28\textwidth]{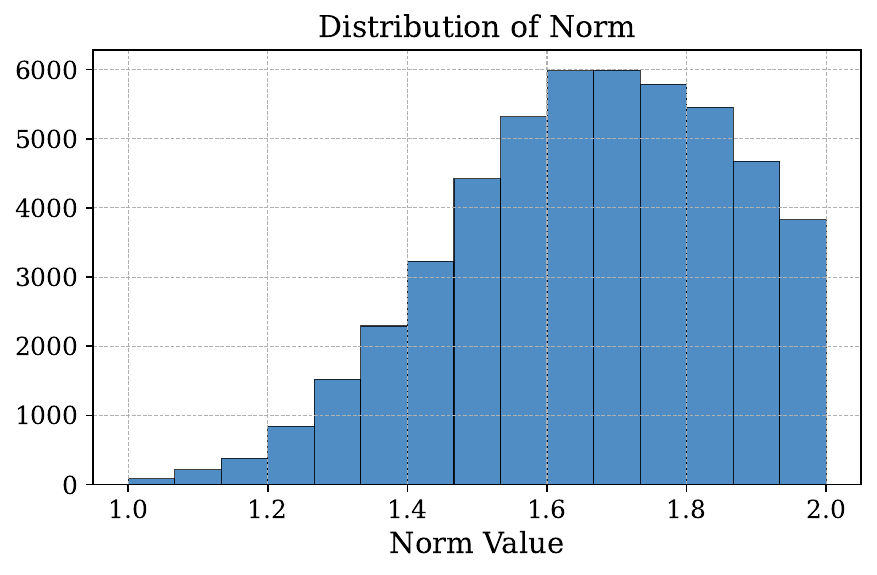}}\quad\\
	\subfloat[][DASTR, $k = 5$. 
	\label{fig:brow_das5_points}]{\includegraphics[width=.28\textwidth]{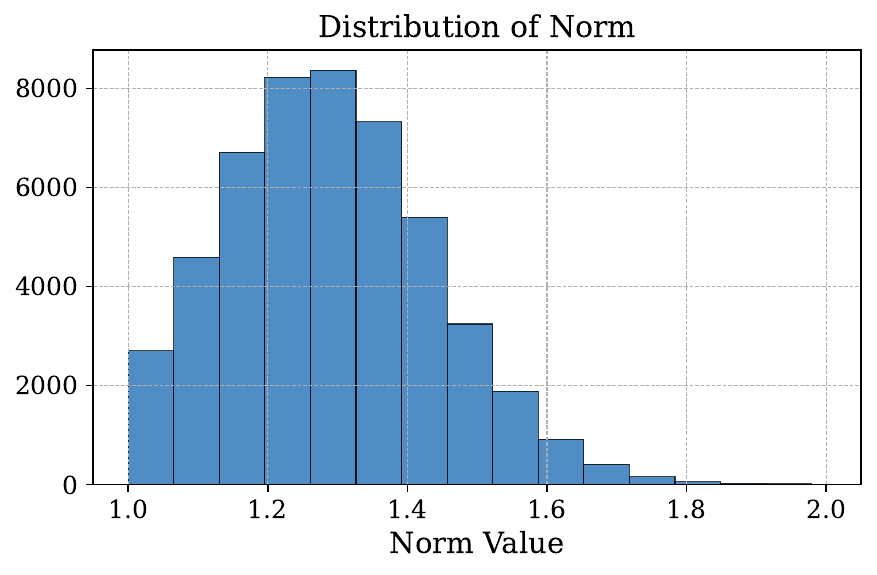}}\quad
	\subfloat[][DASTR, $k = 15$. 
	\label{fig:brow_das15_points}]{\includegraphics[width=.28\textwidth]{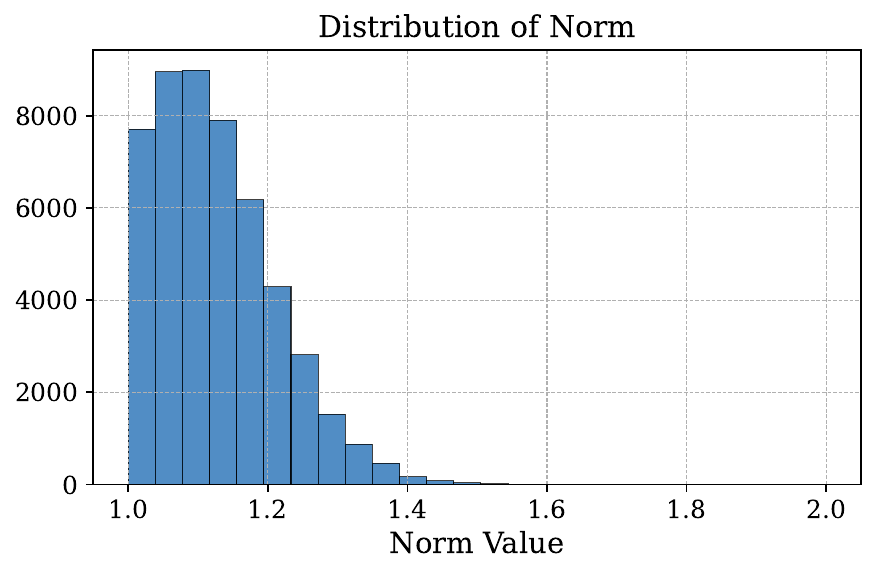}}\quad
	\subfloat[][DASTR, $k = 30$. 
	\label{fig:brow_das30_points}]{\includegraphics[width=.28\textwidth]{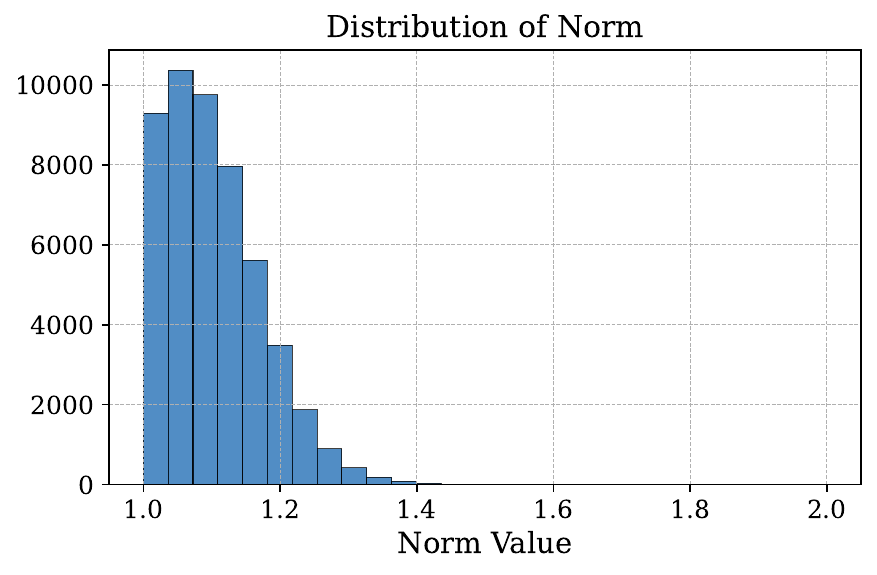}}\\
	\caption{Histogram of the norm of samples, $20$-dimensional test problem. }
	\label{fig:his_norm_sample}
\end{figure}
\begin{table}[!htb]
	\scriptsize
	\centering
	\caption{$20$-dimensional standard Brownian motion test problem: errors for different settings of neural networks and sampling strategies. We take $4$ independent runs to compute the statistics of the error (mean $\pm$ standard deviation).}
	\renewcommand{\arraystretch}{1.5}  %
	\begin{tabular}{ccccc}
		\toprule 
		& & \multicolumn{3}{c}{\textbf{Number of Neurons in Hidden Layer}} \\  
		\hline
		\textbf{Sampling Method} & $\vert \mathsf{S} \vert$ & \textbf{20} & \textbf{50} & \textbf{100} \\ 
		\hline
		\multirow{4}{*}{Uniform} & $5 \times 10^3$ & 0.1767 $\pm$ 0.0240 &  0.1906 $\pm$ 0.0214 & 0.4555 $\pm$ 0.0557  \\
		& $1 \times 10^4$ & 0.1861 $\pm$ 0.0319 &  0.1760 $\pm$ 0.0492 & 0.1310 $\pm$ 0.0197  \\
		& $1.5 \times 10^4$ & 0.2125 $\pm$ 0.0220 &  0.2003 $\pm$ 0.0295  & 0.1454 $\pm$ 0.0609 \\
		& $2 \times 10^4$ & 0.1963 $\pm$ 0.0866 &  0.1611 $\pm$ 0.0227 & 0.1402 $\pm$ 0.0515  \\
		\hline
		\multirow{4}{*}{SDE} &  $5 \times 10^3$ & 0.2127 $\pm$ 0.0802 & 0.2641 $\pm$ 0.0416 & 0.3696 $\pm$ 0.0633 \\
		&  $1 \times 10^4$ &0.2846 $\pm$ 0.0523 & 0.2606 $\pm$ 0.0343 & 0.1586 $\pm$ 0.0179  \\
		&  $1.5 \times 10^4$ &0.2861 $\pm$ 0.0177 & 0.1865 $\pm$ 0.0220 & 0.1706 $\pm$ 0.0434 \\
		&  $2 \times 10^4$ &0.2321 $\pm$ 0.0278 & 0.1864 $\pm$ 0.0254 & 0.1342 $\pm$ 0.0434 \\
		\hline
		\multirow{4}{*}{DASTR (this work)} & $5 \times 10^3$ &0.0996 $\pm$ 0.0374 & 0.1073 $\pm$ 0.0128 & 0.0266 $\pm$ 0.1396  \\
		&  $1 \times 10^4$ &0.0835 $\pm$ 0.0215 & 0.0415 $\pm$ 0.0167 & 0.0410 $\pm$ 0.0106  \\
		&  $1.5 \times 10^4$ &0.0824 $\pm$ 0.0412 & 0.0197 $\pm$ 0.0045& 0.0141 $\pm$ 0.0053   \\
		&  $2 \times 10^4$ &0.0227 $\pm$ 0.0051 & 0.0209 $\pm$ 0.0096  & 0.0114 $\pm$ 0.0021 \\
		\bottomrule
	\end{tabular}
	\label{tab:brow_error}
\end{table}

To see why DASTR works well, let us visualize the $L^2$-norm of samples from different sampling strategies. Figure~\ref{fig:his_norm_sample} shows the histogram of the norm of samples for different sampling strategies. From Figure~\ref{fig:brow_uniform_points} and Figure~\ref{fig:brow_sde_points}, we can see that most of the samples fall into the interval where the norm of samples is near $2$. This means that it is difficult to generate samples in the transition state region using the uniform sampling strategy or SDE. Indeed, in high-dimensional spaces, most of the volume of an object concentrates around its surface \citep{vershynin2018high,wright2021high}. Hence, using uniform samples or samples generated by SDE is inefficient for estimating the committor function. Figures~\ref{fig:brow_das2_points}, \ref{fig:brow_das5_points}, \ref{fig:brow_das15_points}, and \ref{fig:brow_das30_points} show the histogram of the norm of samples from DASTR. These histograms imply that the samples from DASTR capture the information of transitions, which improves the accuracy of estimating the committor function. 
In Table~\ref{tab:brow_error}, we again present the $L^{2}$ relative errors of neural networks with varying numbers of neurons subject to different sample sizes. Our DASTR method is one order of magnitude more accurate than the baseline methods in most settings.

\subsection{Alanine Dipeptide}\label{sec_numexp_ad}
In this test problem, the
isomerization process of the alanine dipeptide in vacuum at $T= 300K$ is studied. This test problem is a benchmark in various literatures \citep{li2019computing, kang2024computing}. There are two parts of this test problem. In section \ref{sec_numexp_ad_explicit_CVs}, we assume that the collective variables are known. Then, the proposed DASTR approach is applied to the collective variables, which will improve the robustness of DASTR in approximating the committor function. In section \ref{sec_numexp_ad_latent_cv}, the collective variables are not explicitly given, which is a more realistic setting. We use an autoencoder to find some latent variables to serve as the collective variables. 

\begin{figure}[!htb]
	\centering \includegraphics[width=0.52\textwidth]{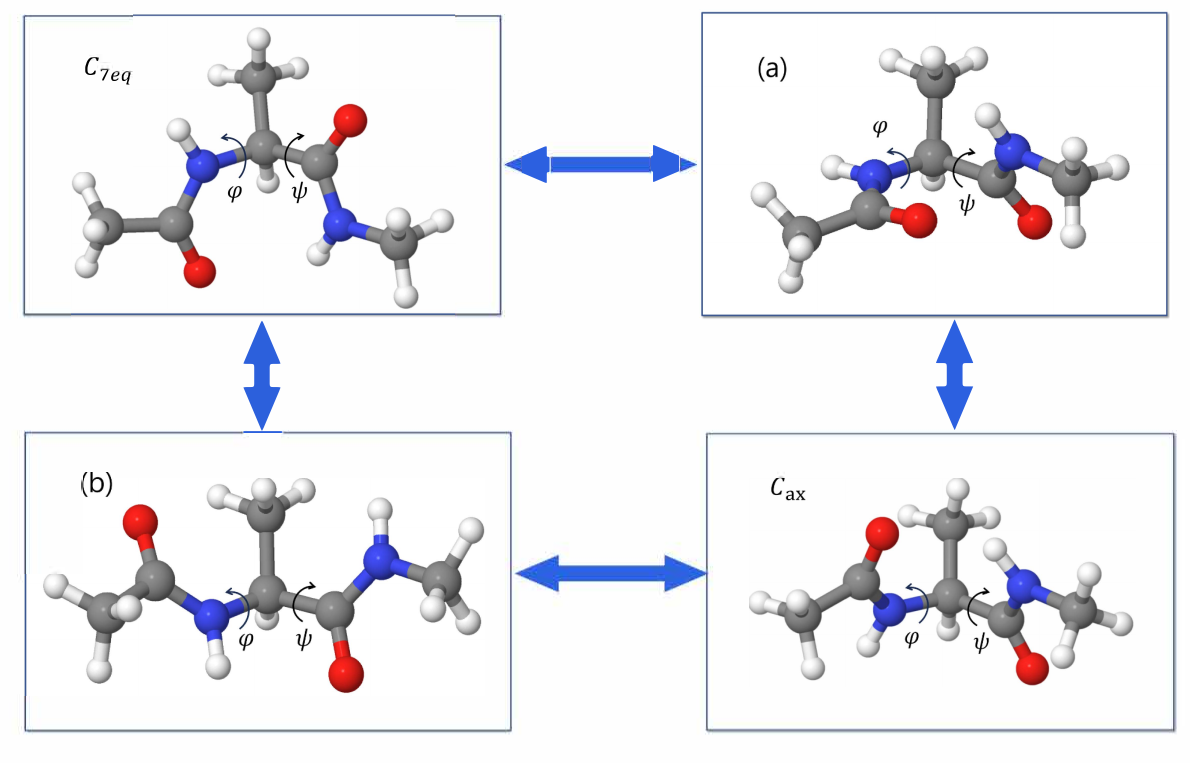}
	\caption{The two metastable states and two transition states of the alanine dipeptide. $C_{7eq}:(\phi, \psi) \approx (-85^\circ, 75^\circ)$ and $C_{ax}:(\phi, \psi) \approx (72^\circ, -75^\circ)$ are two metastable states, $(a):(\phi, \psi) \approx (0^\circ, -65^\circ)$ and $(b):(\phi, \psi) \approx (130^\circ, -125^\circ)$ are two transition states.}\label{fig:ad_structure}
\end{figure}

The molecule we consider here consists of 22 atoms, each of which has three coordinates. This means that the dimension of the state variable is $d = 66$ in \eqref{eq_committor_pde}. There are two important dihedrals related to their configurations: $\phi$ (C-N-CA-C) and $\psi$ (N-CA-C-N). 
The two metastable conformers of the molecule are $C_{7eq}$ and $C_{ax}$ located around $(\phi, \psi) = (-85^\circ, 75^\circ)$ and $(72^\circ, -75^\circ)$ respectively. More specifically, the two metastable sets $A$ and $B$ are defined as \citep{li2019computing}: 
\begin{align*}
	A &= \left\{ \mb{x}: \norm{(\phi(\mb{x}), \psi(\mb{x})) - C_{7eq}}{2} < 10^\circ \right\}, \\
	B &= \left\{ \mb{x}: \norm{(\phi(\mb{x}), \psi(\mb{x})) - C_{ax}}{2} < 10^\circ \right\}.
\end{align*}
In Figure \ref{fig:ad_structure}, the molecule structures of two metastable states and two transition states are displayed. 

The goal is to compute the committor function under the CHARMM force filed \citep{jo2008charmm, brooks2009charmm, lee2016charmm}. Due to the high energy barrier between the two metastable states $A$ and $B$, it is almost impossible for the molecule to cross this barrier from $A$ to $B$. Consequently, sampling in the transition state region with SDE is extremely challenging.

\subsubsection{DASTR with Explicit Collective Variables}\label{sec_numexp_ad_explicit_CVs}
In this part, we study the performance of DASTR with explicit collective variables. The collective variables is set to the two dihedrals $\phi$ (C-N-CA-C) and $\psi$ (N-CA-C-N). For this realistic problem, we need to ensure that the samples from deep generative models obey the molecular configuration, which makes this problem much more challenging to solve. To handle this difficulty, we combine our DASTR method with the umbrella sampling method \citep{kastner2011umbrella} and the collective variables method. Simply speaking, we use the proposed DASTR method to generate the target collective variables in the umbrella potential. The details of the overall procedure can be found in \ref{sec_appendix_setting_ad} and \ref{sec_appendix_us}.

For this problem, it is intractable to obtain the reference solution with grid-based numerical methods. 
To assess the performance of our method, we again consider those samples from the $1/2$-isosurface. More specifically, we first use umbrella sampling (see \ref{sec_appendix_us}) to sample $1 \times 10^{7}$ points. After that, we use the trained model to compute $q_{\mb{\theta}}$ at these sample points and filter to keep points on the set $\Gamma := \{\mb{x}: | q_{\mb{\theta}}(\mb{x})- 0.5 |\} \leq 5 \times 10^{-5}$. We conduct $200$ simulations of SDE for each point in $\Gamma$ to obtain the corresponding trajectory. By counting the number of times of these points first reaching $B$ before $A$, we can estimate $q$ for such points by the definition of committor functions. If the trained model $q_{\mb{\theta}}$ is indeed a good approximation of the committor function, then the probability distribution (in fact, we use the relative frequency to represent the true probability) of reaching $B$ before $A$ should be close to a normal distribution with mean $0.5$ \citep{chen2023committor}.

\begin{figure}[!htb]
	\centering
	\subfloat[][DASTR, $k = 2$. \label{fig:ad_sample_dastr3}]
	{\includegraphics[width=.28\textwidth]{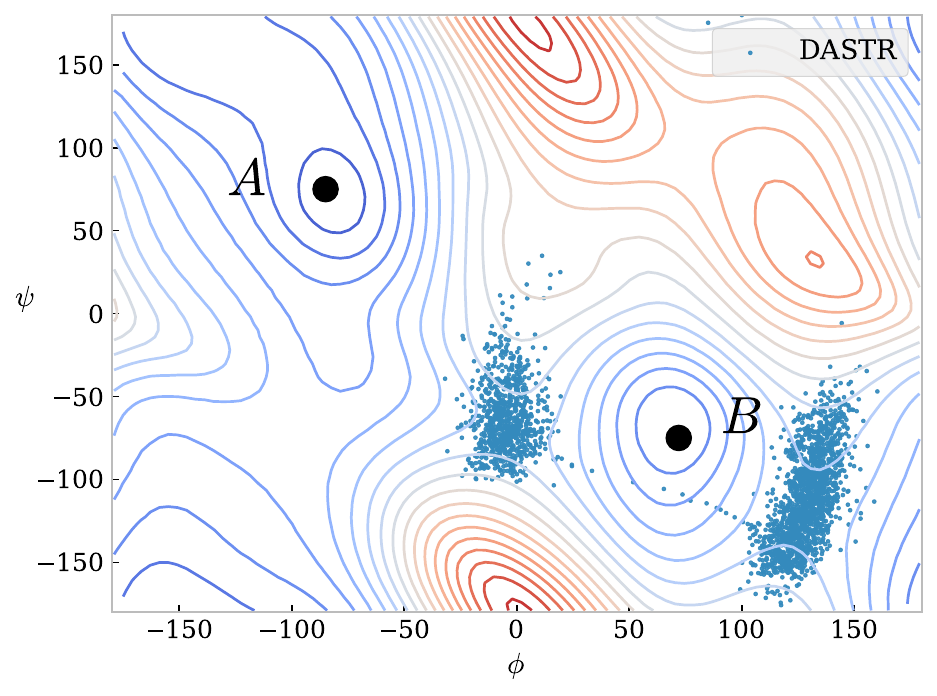}}\quad
	\subfloat[][DASTR, $k = 5$. \label{fig:ad_sample_dastr6}]{\includegraphics[width=.28\textwidth]{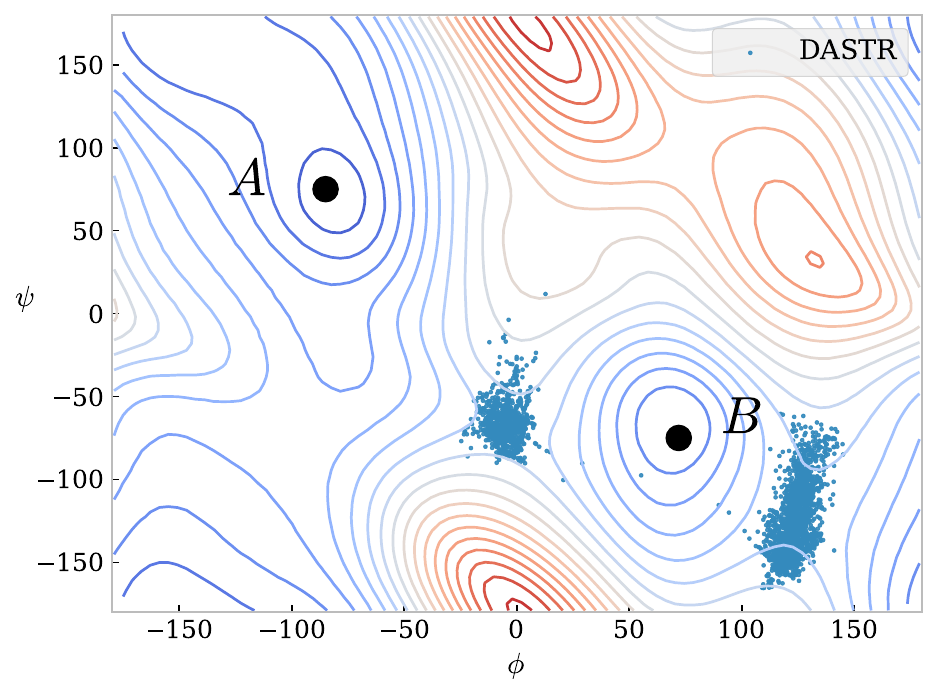}}\quad
	\subfloat[][DASTR, $k = 8$. \label{fig:ad_sample_dastr9}]{\includegraphics[width=.28\textwidth]{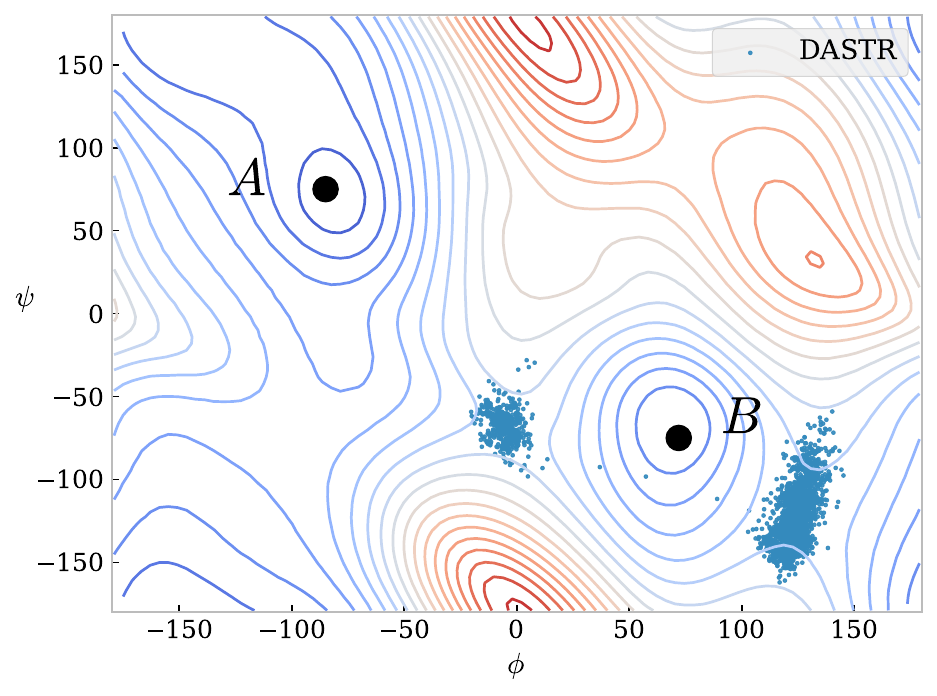}}\\
	\vspace{0.5em}
	\subfloat[][Umbrella sampling, $k = 2$. \label{fig:ad_sample_us3}]{\includegraphics[width=.28\textwidth]{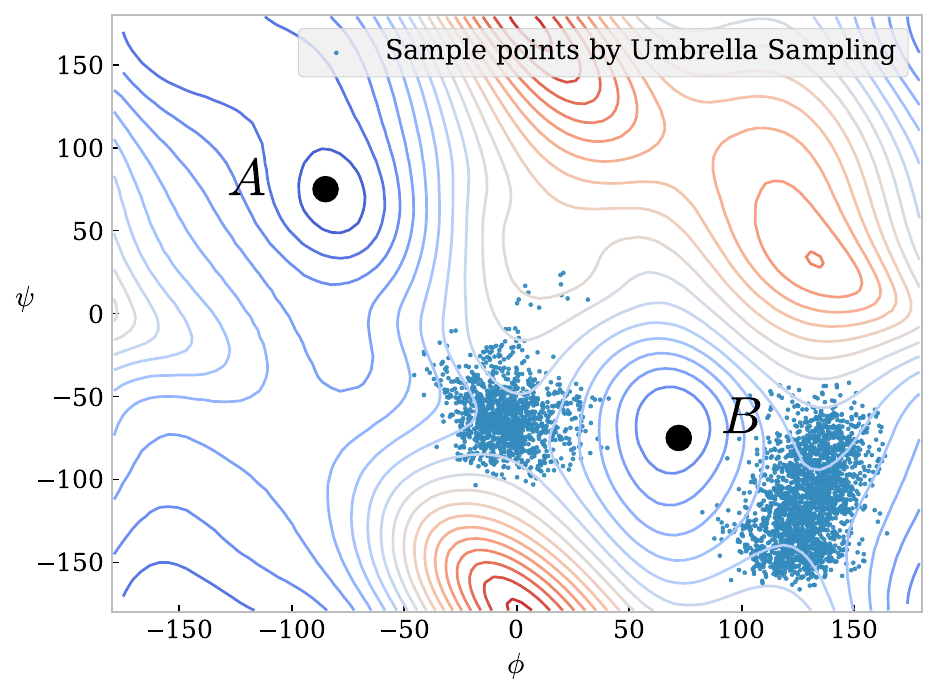}}\quad
	\subfloat[][Umbrella sampling, $k = 5$. \label{fig:ad_sample_us6}]{\includegraphics[width=.28\textwidth]{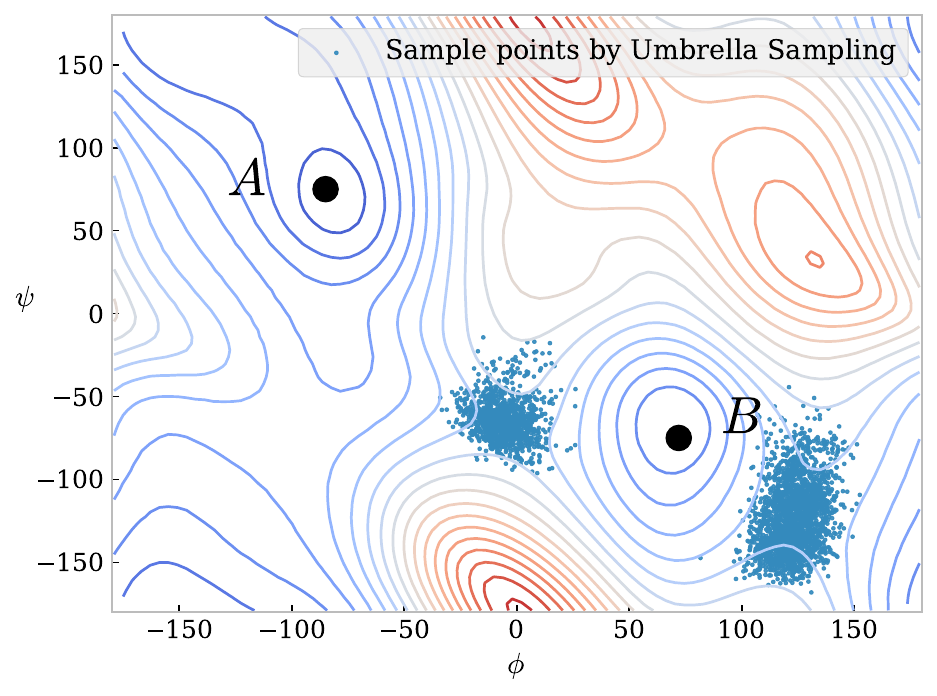}}\quad
	\subfloat[][Umbrella sampling, $k = 8$. \label{fig:ad_sample_us9}]{\includegraphics[width=.28\textwidth]{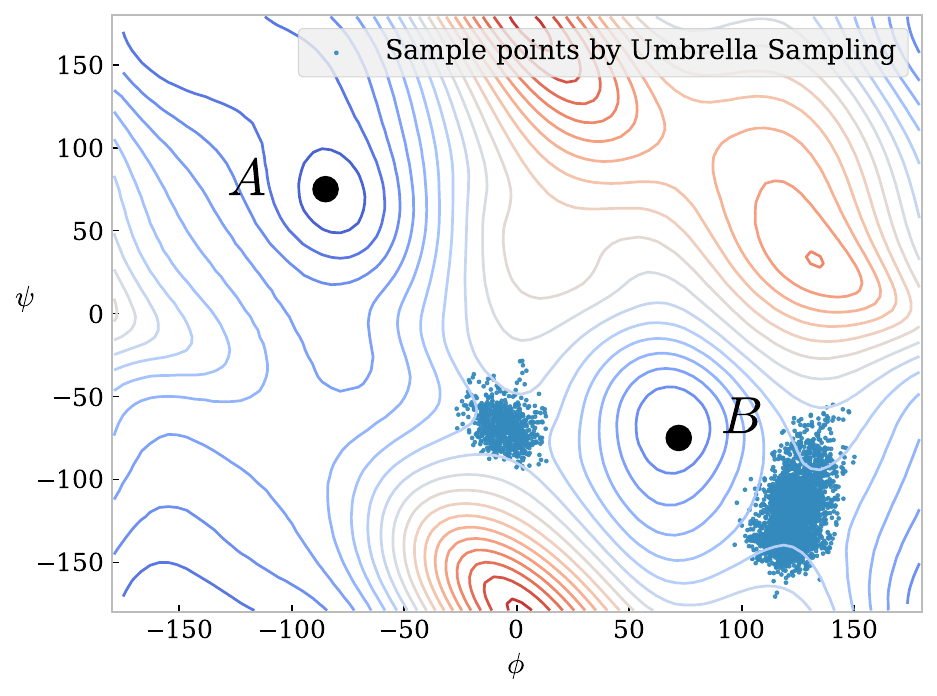}}\\
	\caption{Samples during training for the alanine dipeptide test problem. We use DASTR to generate target CVs in the transition state region; the umbrella sampling method is employed to generate samples around the target CVs to refine the training set. The figures are shown that the samples (scatter plot) distributed on the energy landscape with respect to $\phi$ and $\psi$. }
	\label{fig:ad_DASTR_US_sample}
\end{figure}
\begin{figure}[!htb]
	\centering
	\subfloat[][Metadynamics-5000 terms, 150 neurons. The histogram includes 61 samples. ]{\includegraphics[width=.3\textwidth]{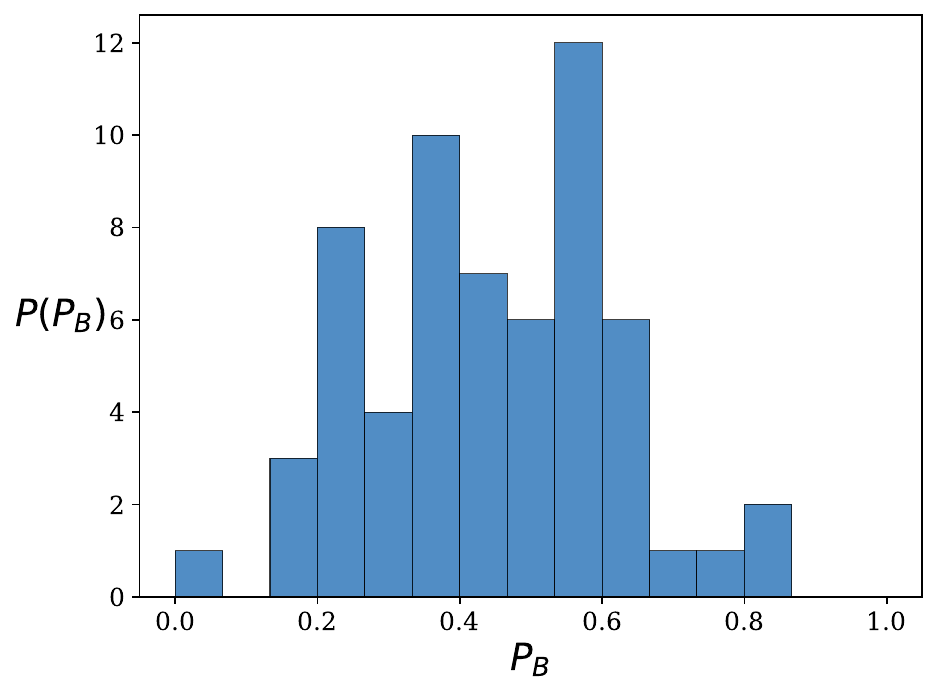}\label{fig:ad_solution_1}}\quad
	\subfloat[][Metadynamics-7500 terms, 150 neurons. The histogram includes 50 samples. ]{\includegraphics[width=.3\textwidth]{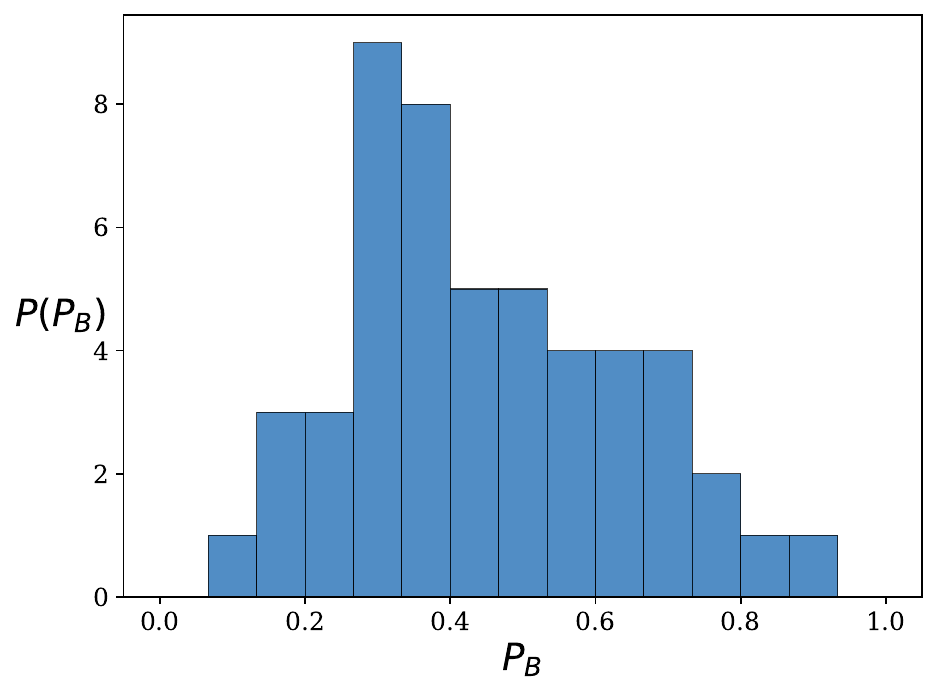}\label{fig:ad_solution_2}}\quad
	\subfloat[][Metadynamics-10000 terms, 150 neurons. The histogram includes 92 samples. ]{\includegraphics[width=.3\textwidth]{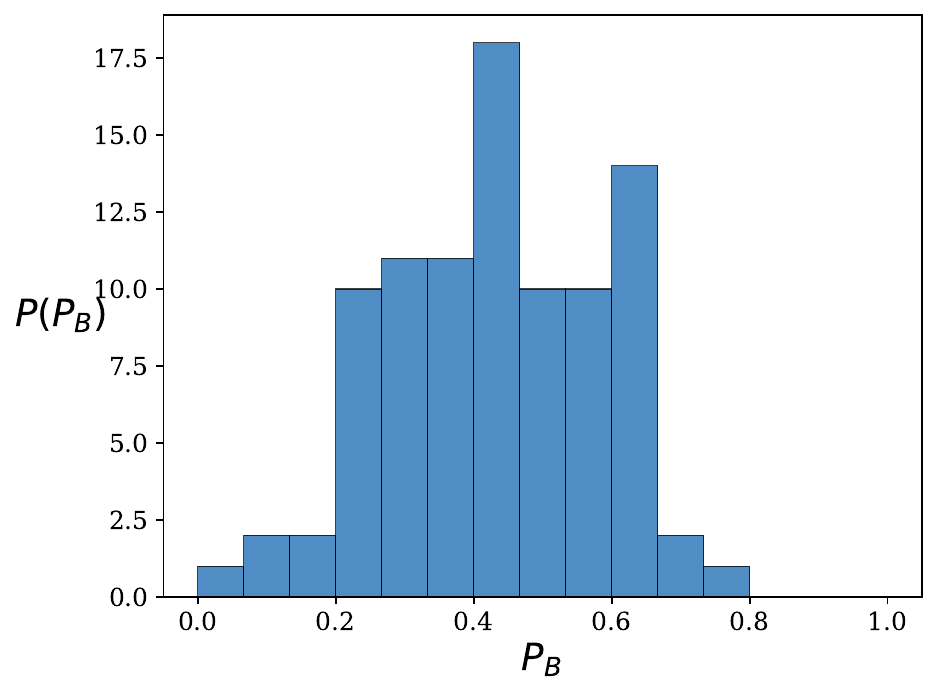}\label{fig:ad_solution_3}} \\
	\vspace{1em}
	\subfloat[][DASTR, 100 neurons. The histogram includes 843 samples.]{\includegraphics[width=.3\textwidth]{ad_solution_DASTR_100_new.pdf}\label{fig:ad_solution_4}}\quad
	\subfloat[][DASTR, 120 neurons. The histogram includes 811 samples.]{\includegraphics[width=.3\textwidth]{ad_solution_DASTR_120_new.pdf}\label{fig:ad_solution_5}}\quad
	\subfloat[][DASTR, 150 neurons. The histogram includes 869 samples.]{\includegraphics[width=.3\textwidth]{ad_solution_DASTR_150_new.pdf}\label{fig:ad_solution_6}}
	\caption{The alanine dipeptide test problem: the histograms of the committor function values on the $1/2$-th isosurface of $q_{\mb{\theta}}$  with different numbers of neurons. $q_{\mb{\theta}}$ is a five-layer fully connected neural network. The training details can be found in \ref{sec_appendix_setting_ad}.}
	\label{fig:ad_solution}
\end{figure}

The results are shown in Figure~\ref{fig:ad_DASTR_US_sample} and Figure~\ref{fig:ad_solution}. In Figure~\ref{fig:ad_sample_dastr3}-\ref{fig:ad_sample_dastr9}, we show the candidate samples generated by DASTR. It is clear that these samples are located in the transition state region. To ensure that the samples obey the molecular configuration, we use the umbrella sampling method to refine them as shown in Figure~\ref{fig:ad_sample_us3}-\ref{fig:ad_sample_us9}. 
From Figure~\ref{fig:ad_solution_1}-\ref{fig:ad_solution_3}, it is seen that the probability distribution is not consistent with a normal distribution with mean 0.5, which means that the trained model using data from metadynamics fails to approximate the committor function near $q \approx 0.5$. Also, the number of points in $\Gamma$ is much smaller than that of DASTR. This is due to the lack of sufficient samples in the transition state region, leading to the large generalization error in this area. In contrast, from Figure~\ref{fig:ad_solution_4}-\ref{fig:ad_solution_6}, it is seen that the approximate committor function values cluster around $1/2$, which indicates that our DASTR method performs significantly better and provides a good approximation on the $1/2$-isosurface.

\subsubsection{DASTR with Latent Collective Variables}\label{sec_numexp_ad_latent_cv}
In the previous experiment, the collective variables $\phi$ and $\psi$ are given. We use KRnet to learn the features of $\phi$ and $\psi$ in the transition state region. Such learned features are used for umbrella sampling to refine the training set. However, this still cannot avoid the need of SDE simulations after training deep generative models. In this part, we use autoencoders to learn the latent collective variables (CVs) that can aid sample generation and avoid repeated simulations of umbrella sampling.

As discussed in section \ref{sec_dastr}, the input of the autoencoder is the coordinates of the 10 heavy atoms of the alanine dipeptide. We perform self-supervised training to train the autoencoder to learn the latent CVs. The KRnet is used to learn the distribution of the latent CVs in the transition state region, which is similar to the approach adopted in section \ref{sec_numexp_ad_explicit_CVs} except for the choice of the latent CVs. The settings of neural networks and training details can be found in \ref{sec_appendix_setting_ad}.

\begin{figure}[!htb]
	\centering
	\subfloat[][$d_{\text{latent}} = 2$. ]{\includegraphics[width=.3\textwidth]{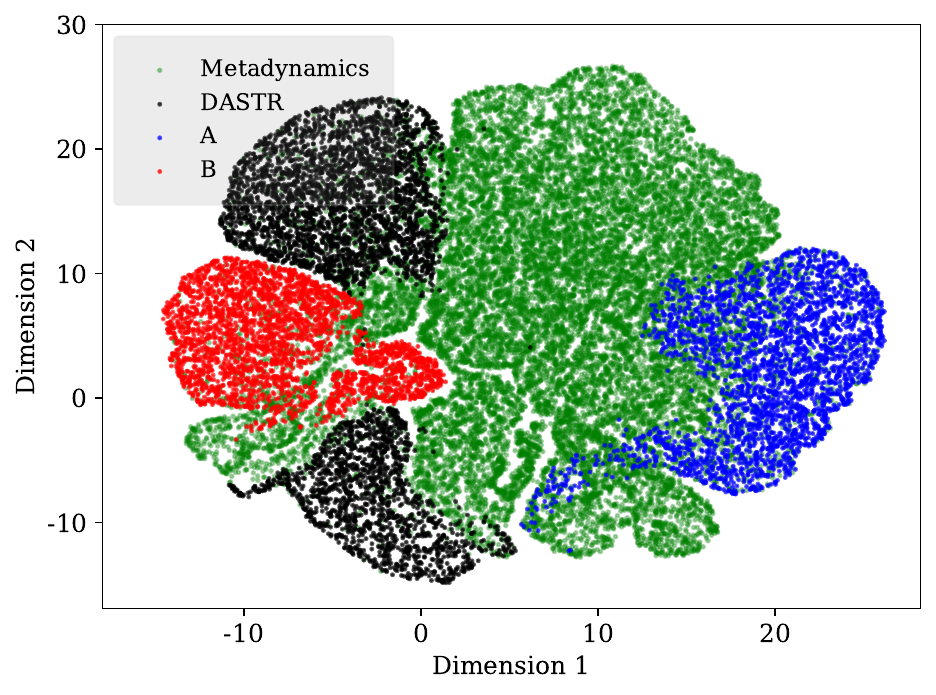}\label{fig:ad_2latent}}\quad
	\subfloat[][$d_{\text{latent}} = 3$. ]{\includegraphics[width=.3\textwidth]{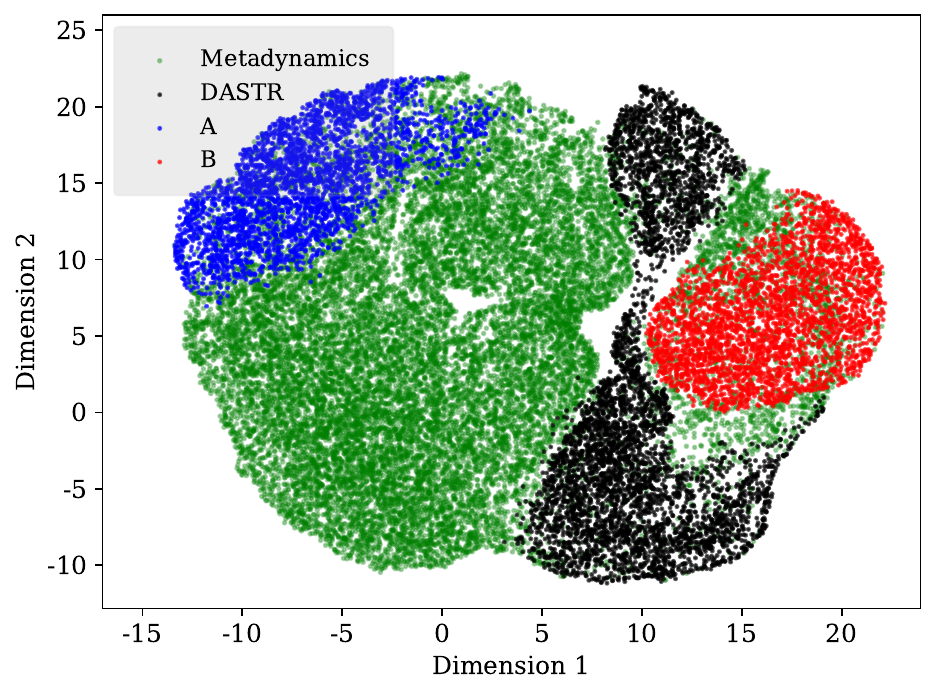}\label{fig:ad_5latent}}\quad
	\subfloat[][$d_{\text{latent}} = 5$. ]{\includegraphics[width=.3\textwidth]{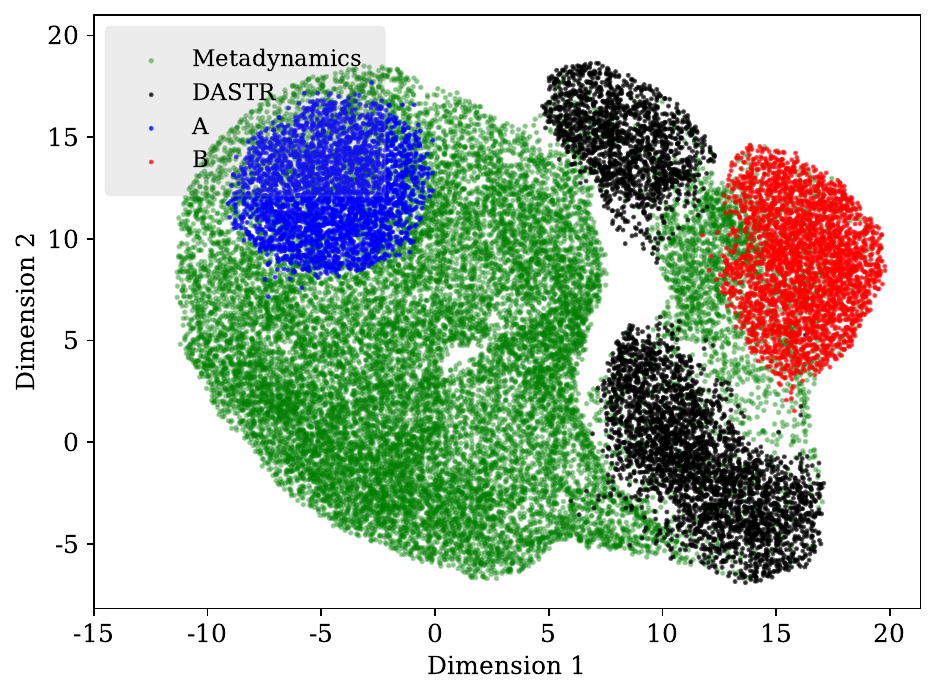}\label{fig:ad_10latent}} 
	\caption{Visualization of the latent collective variables, the two metastable states $A$ and $B$, and samples from DASTR at the final stage in the latent space. The data points are projected onto a two-dimensional plane by UMAP \citep{mcinnes2018umap} for visualization.}
	\label{fig:ad_latent}
\end{figure}

In this experiment, we test three different latent dimensions $d_{\text{latent}} = 2, 3, 5$. In Figure~\ref{fig:ad_latent}, we use UMAP \citep{mcinnes2018umap} to project the data points onto a two-dimensional plane for visualization, where the points include the two metastable states $A$ and $B$, samples from metadynamics, and the latent variables from DASTR at the final stage.

\begin{figure}[!htb]
	\centering
	\subfloat[][The proportion of valid molecular configurations when using the coordinates of the heavy atoms as the input to KRnet.   \label{fig:proportion_valid_configs_cdn}]{\includegraphics[width=.47\textwidth]{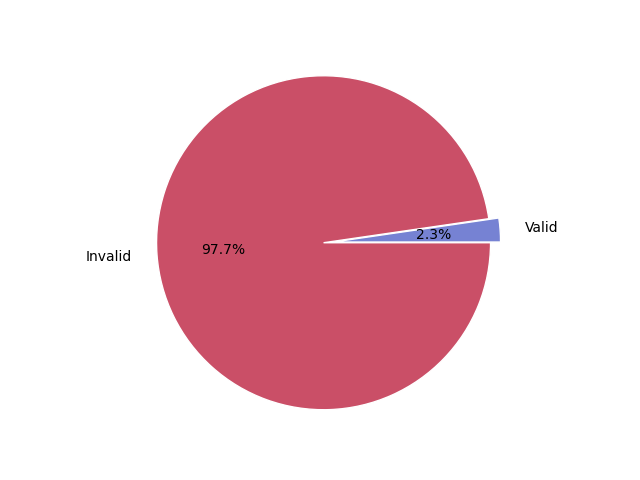}}	\hspace{2em}
    \subfloat[][The proportion of valid molecular configurations when using the latent CVs as the input to KRnet ($d_{\text{latent}} = 3$). \label{fig:proportion_valid_configs_latent_cv}]{\includegraphics[width=.47\textwidth]{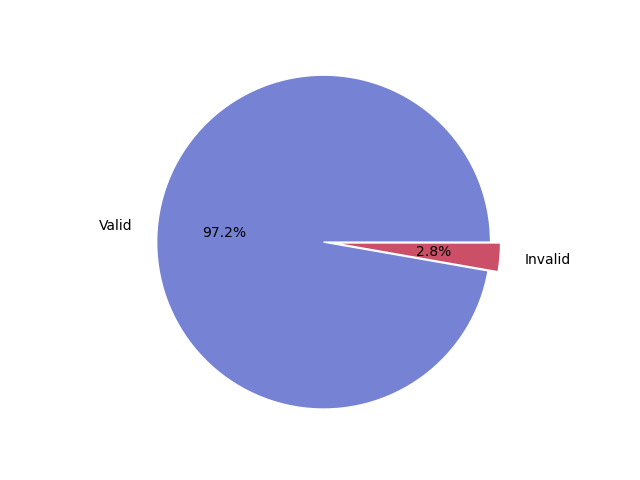}}
	\caption{The proportions of valid molecular configurations for two different settings in DASTR are shown. This figure demonstrates the advantage of performing DASTR in the latent space.}
    \label{fig:comparison_pct_valid_configurations}
\end{figure}

During the adaptive sampling procedure, we need to filter out those samples with excessively high potential energies. This will help avoid generating unreasonable molecular configurations. To this end, we set an energy threshold at 150 kJ/mol in this experiment. This means that any molecules with potential energies exceeding this threshold are discarded when generating new molecules during the adaptive sampling procedure. As a reference, we employ the umbrella sampling method in section \ref{sec_numexp_ad_explicit_CVs} to sample $1\times 10^5$ points in the transition state region, yielding a maximum energy of approximately 115.5 kJ/mol. We generate $1\times 10^5$ samples in the latent space and use the decoder to reconstruct the coordinates of the heavy atoms. The configuration can be completed after adding the hydrogen atoms by PyMOL \citep{schrodinger2015pymol}. For different latent dimensions $d_{\text{latent}} = 2, 3, 5$, the proportions of the samples with energies of less than 150 kJ/mol are approximately $97.52\%, 97.20\%$, and $97.49\%$ respectively. 
For comparison, we also train KRnet using the coordinates of the heavy atoms as the input, and then added hydrogen atoms using PyMOL. In this setting, about $2.3\%$ of the samples have energies of less than 3000 kJ/mol---most of the samples do not have physically reasonable configurations! Figure~\ref{fig:comparison_pct_valid_configurations} shows the comparison of proportions of valid molecular configurations between the vanilla DASTR and the DASTR in the latent space. It is clear that the sampling efficiency is improved significantly when applying DASTR in the latent space. 

\begin{figure}[!htb]
	\centering
	\subfloat[][$d_{\text{latent}} = 2$, $k = 2$. \label{fig:ad_sample_das_3_latent2}]
	{\includegraphics[width=.28\textwidth]{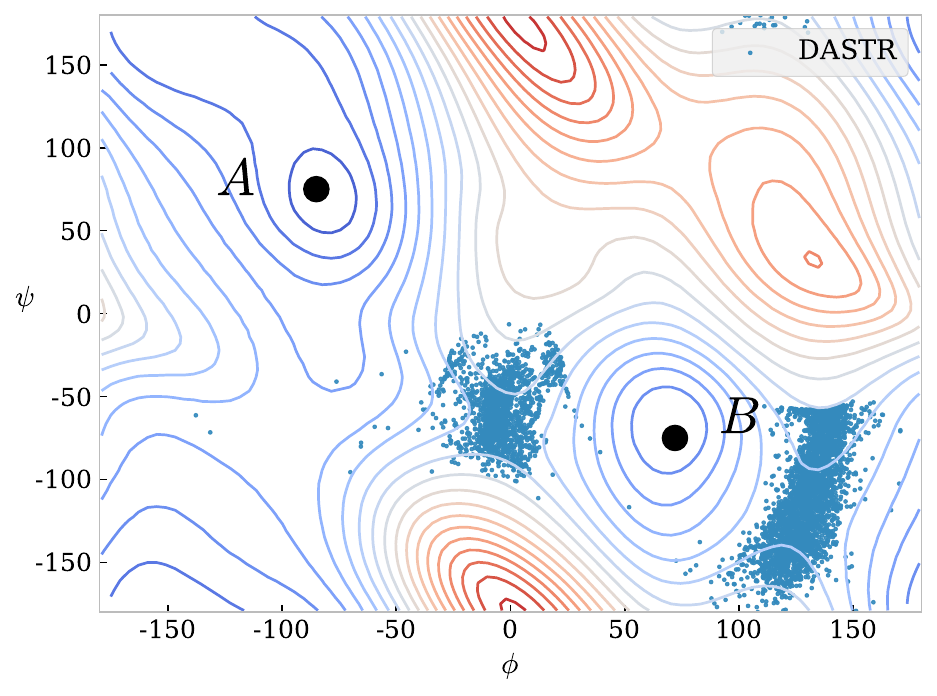}}\quad
	\subfloat[][$d_{\text{latent}} = 2$, $k = 5$. \label{fig:ad_sample_das_6_latent2}]{\includegraphics[width=.28\textwidth]{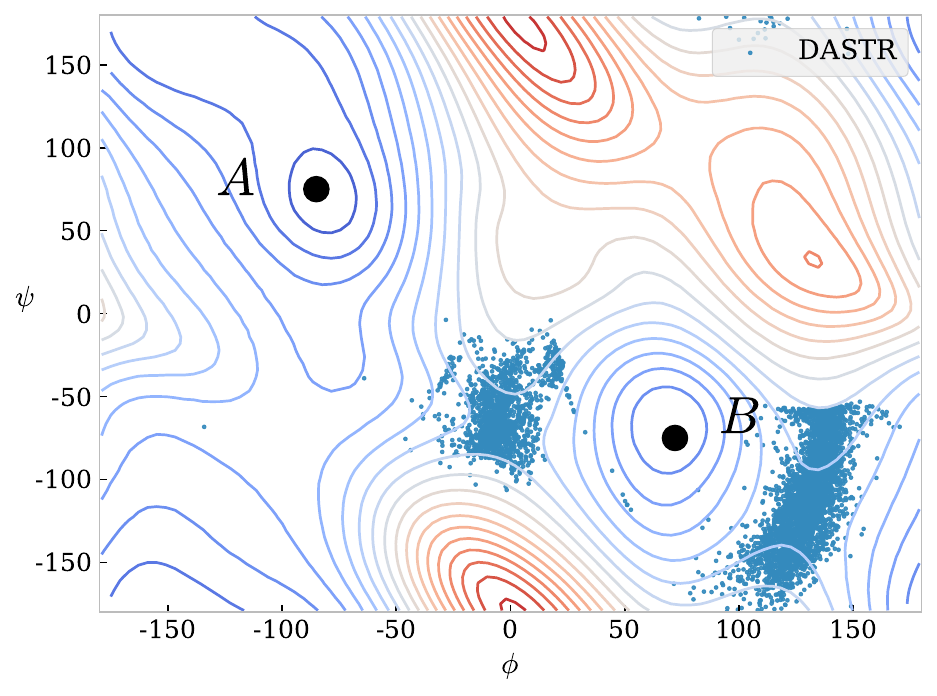}}\quad
	\subfloat[][$d_{\text{latent}} = 2$, $k = 8$. \label{fig:ad_sample_das_9_latent2}]{\includegraphics[width=.28\textwidth]{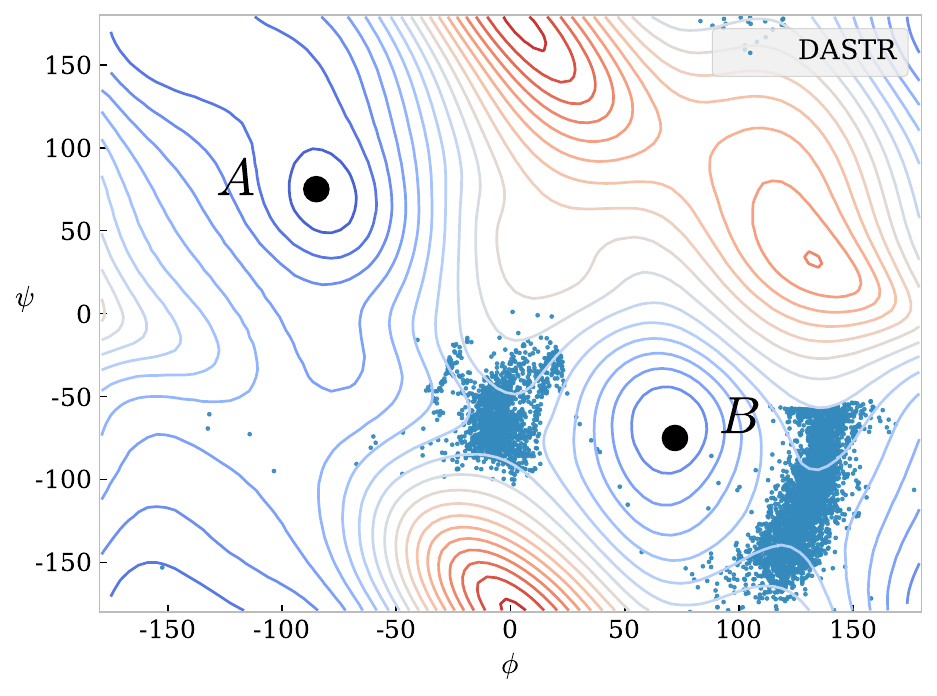}}\\
	\vspace{0.5em}
	\subfloat[][$d_{\text{latent}} = 3$, $k = 2$. \label{fig:ad_sample_das_3_latent3.pdf}]{\includegraphics[width=.28\textwidth]{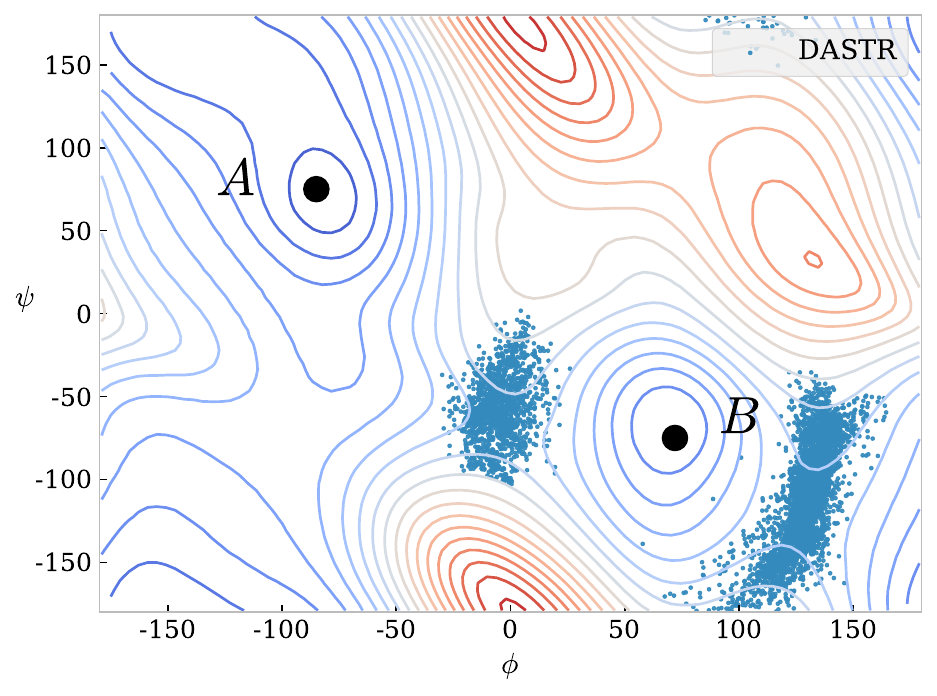}}\quad
	\subfloat[][$d_{\text{latent}} = 3$, $k = 5$. \label{fig:ad_sample_das_6_latent3}]{\includegraphics[width=.28\textwidth]{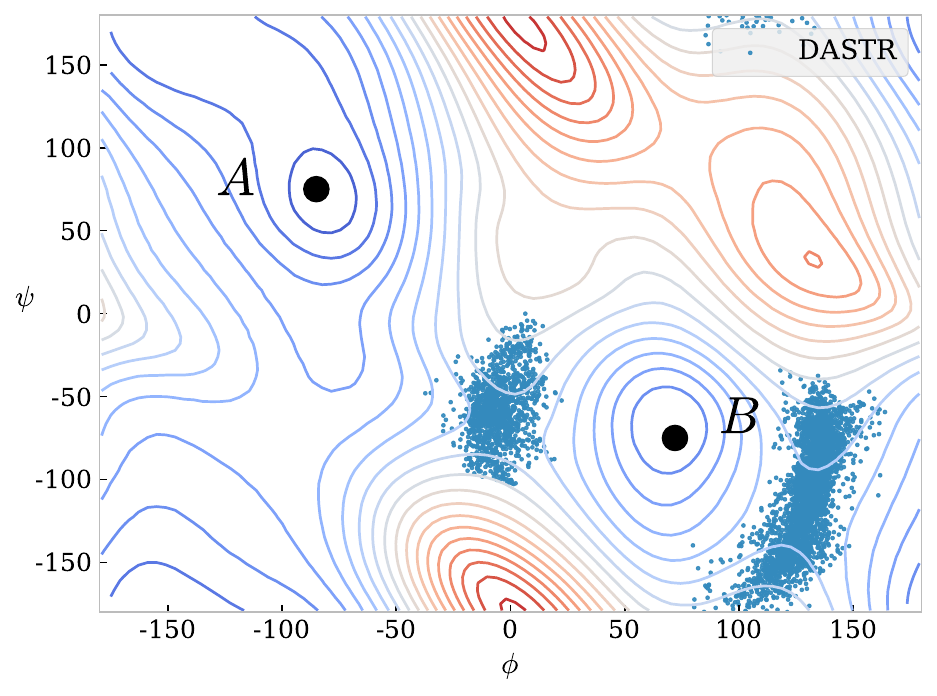}}\quad
	\subfloat[][$d_{\text{latent}} = 3$, $k = 8$. \label{fig:ad_sample_das_9_latent3}]{\includegraphics[width=.28\textwidth]{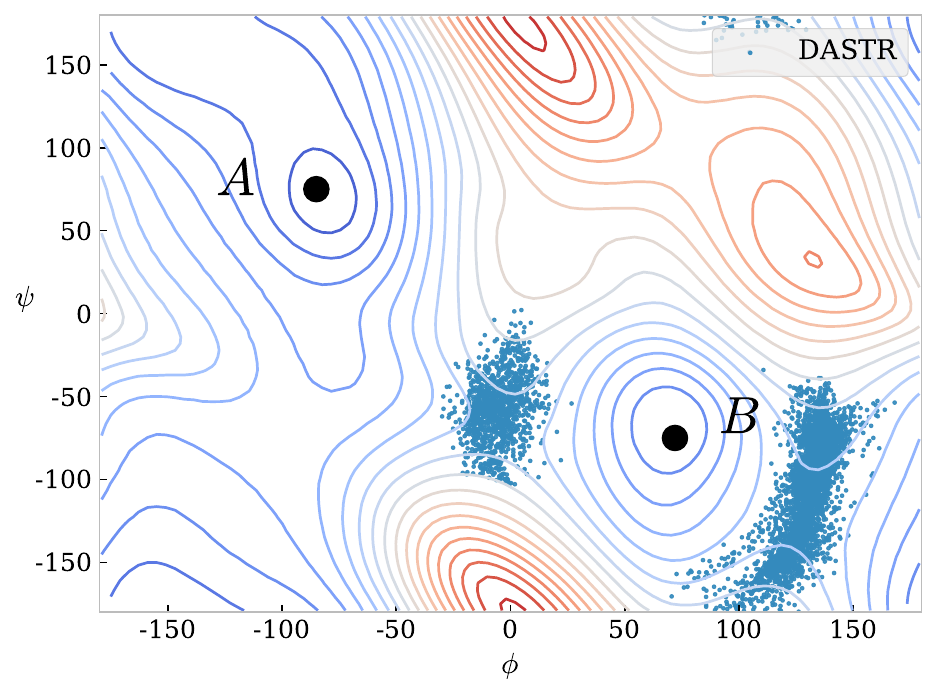}}\\
	\vspace{0.5em}
	\subfloat[][$d_{\text{latent}} = 5$, $k = 2$. \label{fig:ad_sample_das_3_latent5.pdf}]{\includegraphics[width=.28\textwidth]{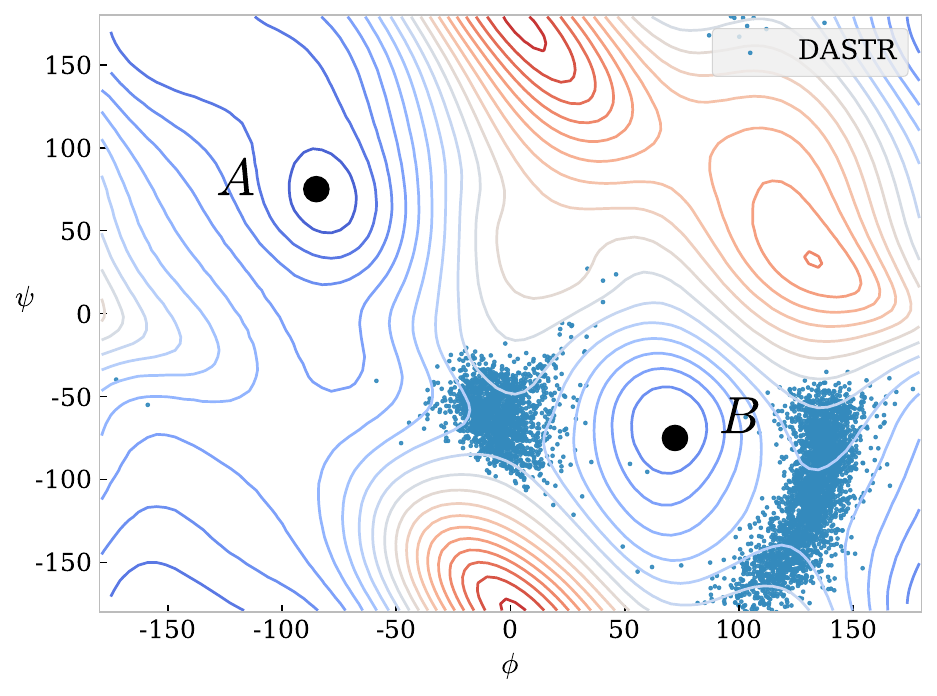}}\quad
	\subfloat[][$d_{\text{latent}} = 5$, $k = 5$. \label{fig:ad_sample_das_6_latent5}]{\includegraphics[width=.28\textwidth]{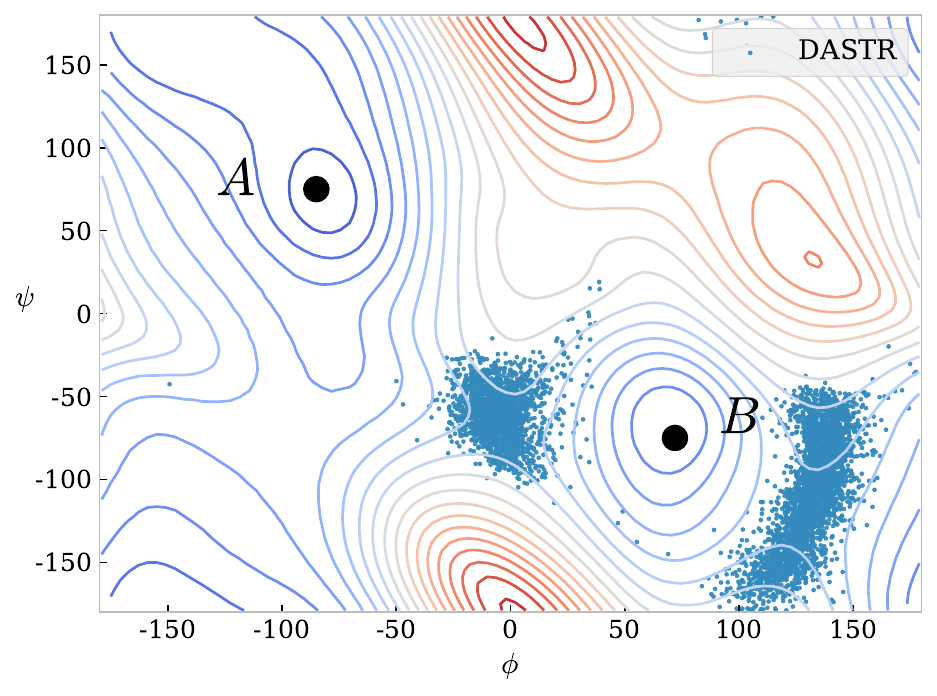}}\quad
	\subfloat[][$d_{\text{latent}} = 5$, $k = 8$. \label{fig:ad_sample_das_9_latent5}]{\includegraphics[width=.28\textwidth]{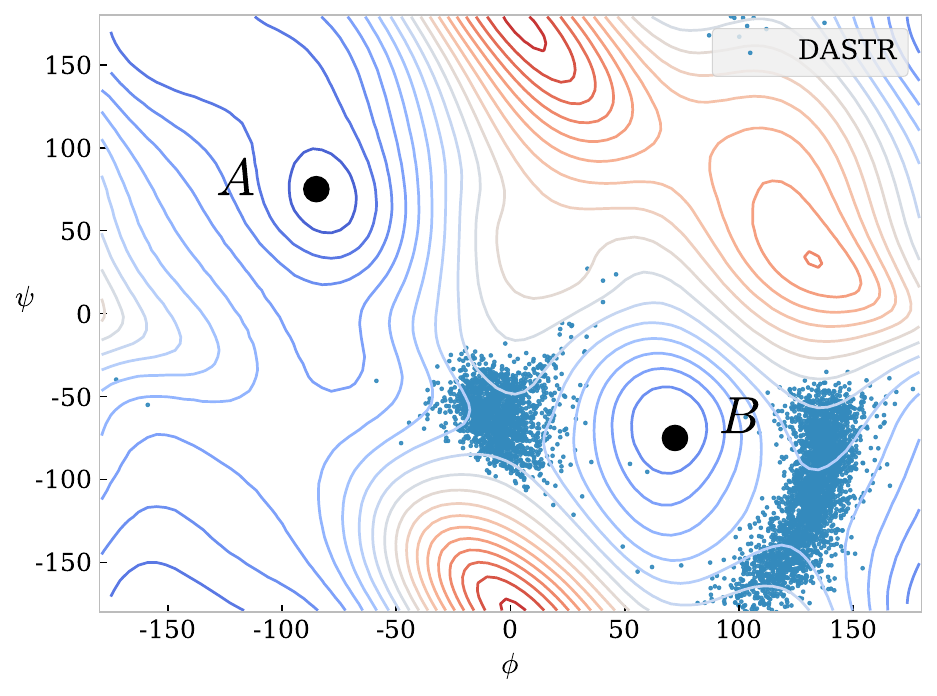}}\\
	\caption{Samples during training for different latent dimensions, the alanine dipeptide test problem. The figures are shown that the samples (scatter plot) distributed on the energy landscape with respect to $\phi$ and $\psi$.}
	\label{fig:ad_DASTR_sample_latent}
\end{figure}

\begin{figure}[!htb]
	\centering
	\subfloat[][$d_{\text{latent}} = 2$. The histogram includes 606 samples. ]{\includegraphics[width=.3\textwidth]{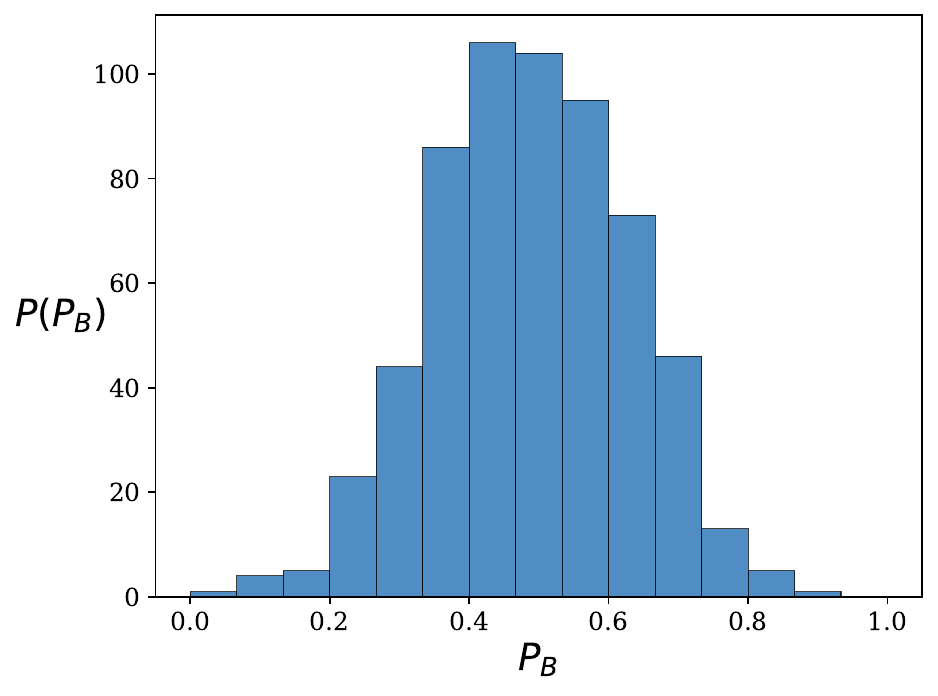}\label{fig:ad_solution_latent2}}\quad
	\subfloat[][$d_{\text{latent}} = 3$. The histogram includes 657 samples. ]{\includegraphics[width=.3\textwidth]{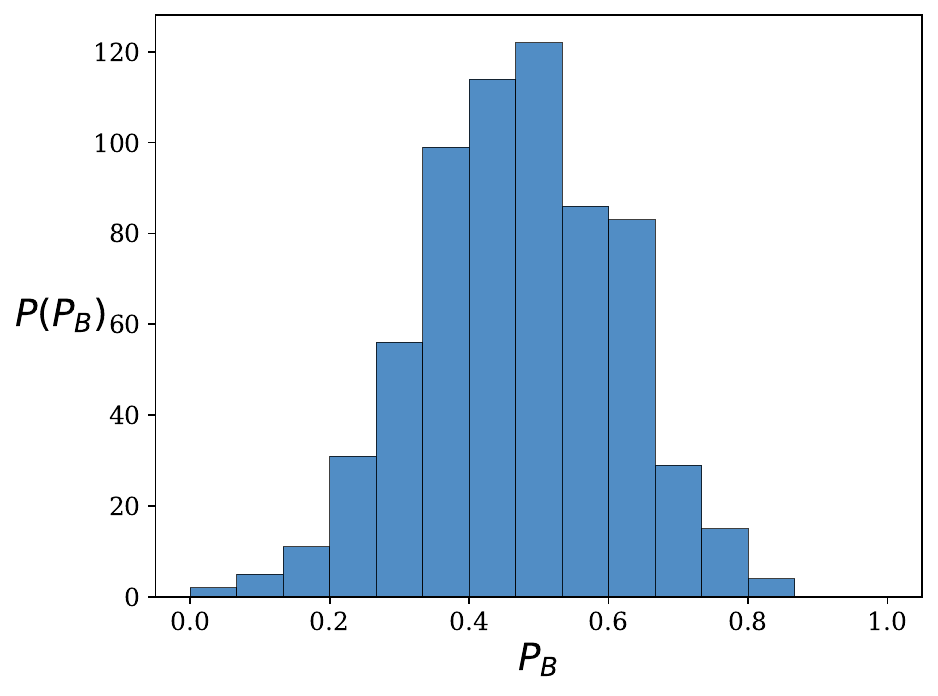}\label{fig:ad_solution_latent3}}\quad
	\subfloat[][$d_{\text{latent}} = 5$. The histogram includes 605 samples.]{\includegraphics[width=.3\textwidth]{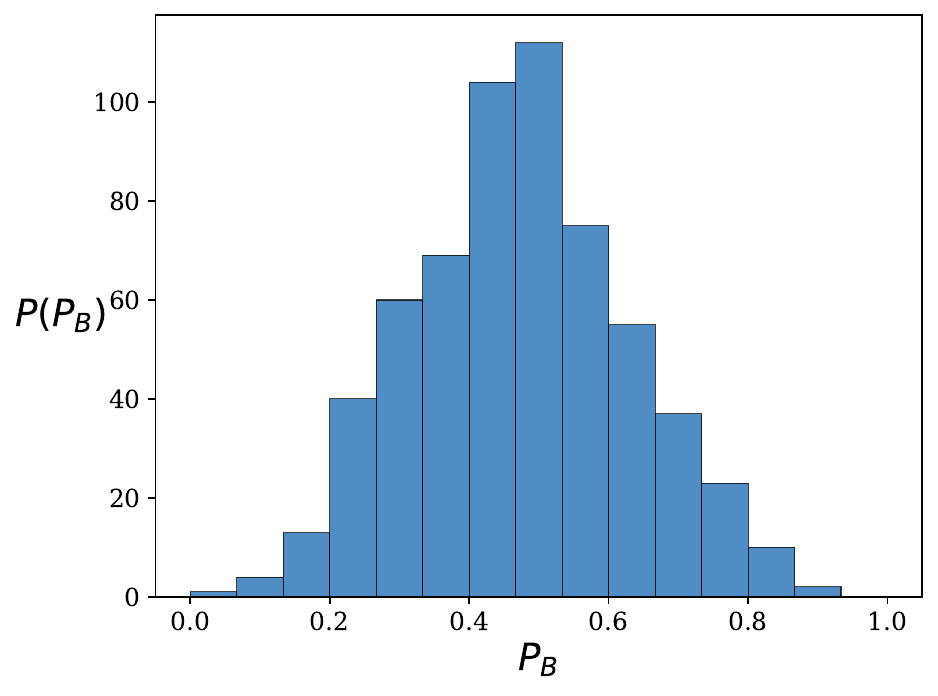}\label{fig:ad_solution_latent5}} \\
	\caption{Conducting DASTR in the latent space for the alanine dipeptide test problem: the histograms of the committor function values on the $1/2$-th isosurface of $q_{\mb{\theta}}$ for different latent dimensions.}
	\label{fig:ad_solution_latent}
\end{figure}

The decoding step requires almost no time when using the autoencoder to generate new molecules. The main time cost for this step is from the hydrogen atom completion in PyMOL, which is also negligible. In Table~\ref{tab:ad_sample_time}, we compare the time cost of conducting DASTR in the latent space and DASTR with umbrella sampling for different numbers of samples. One can observe that the time required to generate the molecules using the latent CVs is less than $4 \%$ of that of the strategy in section \ref{sec_numexp_ad_explicit_CVs}. With the autoencoder, one can apply the proposed DASTR method to the latent space. This technique eliminates the need for simulating SDE to obtain samples in the transition state region and significantly reduces the computational cost, as demonstrated in Table~\ref{tab:ad_sample_time}.
The results are shown in Figure~\ref{fig:ad_DASTR_sample_latent} and Figure~\ref{fig:ad_solution_latent}. As shown in Figure~\ref{fig:ad_DASTR_sample_latent}, it is clear that these samples are located in the transition state region for different latent dimensions studied. From Figure\ref{fig:ad_solution} and Figure\ref{fig:ad_solution_latent}, it is evident that the latter has a smaller variance and thus has a better approximation of the committor function on the $1/2$-isosurface.

\begin{table}[!htb]
	\scriptsize
	\centering
	\caption{Time comparison of DASTR with the explicit collective variables and umbrella sampling and DASTR with the learned latent variables for different numbers of samples (the unit is seconds).}
	\setlength{\tabcolsep}{14pt}  
	\renewcommand{\arraystretch}{1.6}  %
	\begin{tabular}{cccccc}
		\toprule 
		& & \multicolumn{3}{c}{\textbf{Number of Samples}} \\  
		\hline
		\textbf{Sampling Method} & $1 \times 10^4 $ & $2 \times 10^4 $ & $5 \times 10^4 $ & $1 \times 10^5 $ &
		$2 \times 10^5 $\\
		\hline
		{DASTR with umbrella sampling} & 234.01 s & 476.19 s &  1213.17 s & 2406.86 s & 4771.42 s \\
		{DASTR with learned latent variables} & 10.26 s & 18.10 s & 46.33 s & 92.94 s & 175.98 s \\
		\bottomrule
	\end{tabular}
	\label{tab:ad_sample_time}
\end{table}

\section{Conclusion}\label{sec_concl}
We have developed a novel deep adaptive sampling approach on rare transition paths (DASTR) for estimating the high-dimensional committor function. With DASTR, the scarcity of effective data points can be addressed, and the performance of neural network approximation for the high-dimensional committor function is improved significantly.

For high-dimensional realistic molecular systems, to address the issue that deep generative models alone may fail to generate physically reasonable molecular configurations, we apply DASTR to the latent space, where two options for selecting the latent variables are provided. The first option is to combine physically explicit collective variables with umbrella sampling, and the second is to train an autoencoder to find the latent collective variables. Compared to the samples from the directly approximated high-dimensional distribution, the two latent-space-based approaches take into account the physics either through domain-specific knowledge or data. Numerical experiments show that the second choice does not require domain-specific knowledge, except for data used to select the collective variables, potentially providing a generic strategy to deal with larger, more realistic molecular systems. Many questions remain open, especially regarding the correlation between representation learning and physically consistent sample generation. These questions will be left for future study.

\section*{Acknowledgements}
K. Tang is partially supported by the Natural Science Foundation of Hunan Province (2024JJ6003). X. Wan has been supported by NSF grant DMS-1913163. C. Yang has been supported by NSFC grant 12131002.



\bibliographystyle{elsarticle-num}


\bibliography{tang.bib}

\begin{thebibliography}{10}
\expandafter\ifx\csname url\endcsname\relax
  \def\url#1{\texttt{#1}}\fi
\expandafter\ifx\csname urlprefix\endcsname\relax\def\urlprefix{URL }\fi
\expandafter\ifx\csname href\endcsname\relax
  \def\href#1#2{#2} \def\path#1{#1}\fi

\bibitem{okuyama1998transition}
N.~Okuyama-Yoshida, M.~Nagaoka, T.~Yamabe, Transition-state optimization on
  free energy surface: {T}oward solution chemical reaction ergodography,
  International Journal of Quantum Chemistry 70~(1) (1998) 95--103.

\bibitem{E2006towards}
W.~E, E.~Vanden-Eijnden, Towards a theory of transition paths, Journal of
  Statistical Physics 123~(3) (2006) 503--523.

\bibitem{berteotti2009protein}
A.~Berteotti, A.~Cavalli, D.~Branduardi, F.~L. Gervasio, M.~Recanatini,
  M.~Parrinello, Protein conformational transitions: the closure mechanism of a
  kinase explored by atomistic simulations, Journal of the American Chemical
  Society 131~(1) (2009) 244--250.

\bibitem{E2010transition}
W.~E, E.~Vanden-Eijnden, Transition-path theory and path-finding algorithms for
  the study of rare events., Annual Review of Physical Chemistry 61 (2010)
  391--420.

\bibitem{lai2018point}
R.~Lai, J.~Lu, Point cloud discretization of {F}okker--{P}lanck operators for
  committor functions, Multiscale Modeling \& Simulation 16~(2) (2018)
  710--726.

\bibitem{li2019computing}
Q.~Li, B.~Lin, W.~Ren, Computing committor functions for the study of rare
  events using deep learning, The Journal of Chemical Physics 151~(5) (2019)
  054112.

\bibitem{chen2023committor}
Y.~Chen, J.~Hoskins, Y.~Khoo, M.~Lindsey, Committor functions via tensor
  networks, Journal of Computational Physics 472 (2023) 111646.

\bibitem{khoo2019solving}
Y.~Khoo, J.~Lu, L.~Ying, Solving for high-dimensional committor functions using
  artificial neural networks, Research in the Mathematical Sciences 6~(1)
  (2019) 1--13.

\bibitem{li2022semigroup}
H.~Li, Y.~Khoo, Y.~Ren, L.~Ying, A semigroup method for high dimensional
  committor functions based on neural network, in: Mathematical and Scientific
  Machine Learning, PMLR, 2022, pp. 598--618.

\bibitem{li2020solving}
H.~Li, Y.~Khoo, Y.~Ren, L.~Ying, Solving for high dimensional committor
  functions using neural network with online approximation to derivatives,
  arXiv preprint arXiv:2012.06727 (2020).

\bibitem{rotskoff2022active}
G.~M. Rotskoff, A.~R. Mitchell, E.~Vanden-Eijnden, Active importance sampling
  for variational objectives dominated by rare events: {C}onsequences for
  optimization and generalization, in: Mathematical and Scientific Machine
  Learning, PMLR, 2022, pp. 757--780.

\bibitem{hasyim2022supervised}
M.~R. Hasyim, C.~H. Batton, K.~K. Mandadapu, Supervised learning and the
  finite-temperature string method for computing committor functions and
  reaction rates, The Journal of Chemical Physics 157~(18) (2022).

\bibitem{kang2024computing}
P.~Kang, E.~Trizio, M.~Parrinello, Computing the committor with the committor
  to study the transition state ensemble, Nature Computational Science (2024)
  1--10.

\bibitem{lin2024deep}
B.~Lin, W.~Ren, Deep learning method for computing committor functions with
  adaptive sampling, arXiv preprint arXiv:2404.06206 (2024).

\bibitem{gao2021active}
W.~Gao, C.~Wang, Active learning based sampling for high-dimensional nonlinear
  partial differential equations, Journal of Computational Physics 475 (2023)
  111848.

\bibitem{tang_das}
K.~Tang, X.~Wan, C.~Yang, D{AS-PINN}s: A deep adaptive sampling method for
  solving high-dimensional partial differential equations, Journal of
  Computational Physics 476 (2023) 111868.

\bibitem{wang_das2}
X.~Wang, K.~Tang, J.~Zhai, X.~Wan, C.~Yang, Deep {A}daptive {S}ampling for
  {S}urrogate {M}odeling {W}ithout {L}abeled {D}ata, Journal of Scientific
  Computing 101~(3) (2024) 77.
\newblock \href {https://doi.org/10.1007/s10915-024-02711-1}
  {\path{doi:10.1007/s10915-024-02711-1}}.

\bibitem{tang_aas}
K.~Tang, J.~Zhai, X.~Wan, C.~Yang, Adversarial adaptive sampling: Unify {PINN}
  and optimal transport for the approximation of {PDE}s, in: The Twelfth
  International Conference on Learning Representations, 2024.

\bibitem{gao2023failure}
Z.~Gao, L.~Yan, T.~Zhou, Failure-informed adaptive sampling for pinns, SIAM
  Journal on Scientific Computing 45~(4) (2023) A1971--A1994.

\bibitem{jiao2023gas}
Y.~Jiao, D.~Li, X.~Lu, J.~Z. Yang, C.~Yuan, A {G}aussian mixture
  distribution-based adaptive sampling method for physics-informed neural
  networks, Engineering Applications of Artificial Intelligence 135 (2024)
  108770.

\bibitem{czibula2021autoppi}
G.~Czibula, A.-I. Albu, M.~I. Bocicor, C.~Chira, Autoppi: An ensemble of deep
  autoencoders for protein--protein interaction prediction, Entropy 23~(6)
  (2021) 643.

\bibitem{alam2019learning}
F.~F. Alam, T.~Rahman, A.~Shehu, Learning reduced latent representations of
  protein structure data, in: Proceedings of the 10th ACM International
  Conference on Bioinformatics, Computational Biology and Health Informatics,
  2019, pp. 592--597.

\bibitem{hawkins2021generating}
A.~Hawkins-Hooker, F.~Depardieu, S.~Baur, G.~Couairon, A.~Chen, D.~Bikard,
  Generating functional protein variants with variational autoencoders, PLoS
  Computational Biology 17~(2) (2021) e1008736.

\bibitem{sirignano2018dgm}
J.~Sirignano, K.~Spiliopoulos, D{GM}: A deep learning algorithm for solving
  partial differential equations, Journal of Computational Physics 375 (2018)
  1339--1364.

\bibitem{raissi2019physics}
M.~Raissi, P.~Perdikaris, G.~E. Karniadakis, Physics-informed neural networks:
  A deep learning framework for solving forward and inverse problems involving
  nonlinear partial differential equations, Journal of Computational Physics
  378 (2019) 686--707.

\bibitem{karniadakis2021physics}
G.~E. Karniadakis, I.~G. Kevrekidis, L.~Lu, P.~Perdikaris, S.~Wang, L.~Yang,
  Physics-informed machine learning, Nature Reviews Physics 3~(6) (2021)
  422--440.

\bibitem{weinan2018deep}
W.~E, B.~Yu, The deep {R}itz method: A deep learning-based numerical algorithm
  for solving variational problems, Communications in Mathematics and
  Statistics 6~(1) (2018) 1--12.

\bibitem{LMDeepNitscheMethod}
Y.~Liao, P.~Ming, Deep {N}itsche {M}ethod: Deep {R}itz {M}ethod with
  {E}ssential {B}oundary {C}onditions, Communications in Computational Physics
  29~(5) (2021) 1365--1384.

\bibitem{lu2021priori}
Y.~Lu, J.~Lu, M.~Wang, A priori generalization analysis of the deep {R}itz
  method for solving high dimensional elliptic partial differential equations,
  in: Conference on Learning Theory, PMLR, 2021, pp. 3196--3241.

\bibitem{kastner2011umbrella}
J.~K{\"a}stner, Umbrella sampling, Wiley Interdisciplinary Reviews:
  Computational Molecular Science 1~(6) (2011) 932--942.

\bibitem{bussi2020using}
G.~Bussi, A.~Laio, Using metadynamics to explore complex free-energy
  landscapes, Nature Reviews Physics 2~(4) (2020) 200--212.

\bibitem{barducci2008well}
A.~Barducci, G.~Bussi, M.~Parrinello, Well-tempered metadynamics: a smoothly
  converging and tunable free-energy method, Physical Review Letters 100~(2)
  (2008) 020603.

\bibitem{dinh2016density}
L.~Dinh, J.~Sohl-Dickstein, S.~Bengio, Density estimation using real {NVP},
  arXiv preprint arXiv:1605.08803 (2016).

\bibitem{kingma2018glow}
D.~P. Kingma, P.~Dhariwal, Glow: Generative flow with invertible 1x1
  convolutions, in: Advances in Neural Information Processing Systems, 2018,
  pp. 10215--10224.

\bibitem{chen2018neural}
T.~Q. Chen, Y.~Rubanova, J.~Bettencourt, D.~K. Duvenaud, Neural ordinary
  differential equations, in: Advances in {N}eural {I}nformation {P}rocessing
  {S}ystems, 2018, pp. 6571--6583.

\bibitem{song2021scorebased}
Y.~Song, J.~Sohl-Dickstein, D.~P. Kingma, A.~Kumar, S.~Ermon, B.~Poole,
  Score-based generative modeling through stochastic differential equations,
  in: International Conference on Learning Representations, 2021.

\bibitem{tangwandensity2020}
K.~Tang, X.~Wan, Q.~Liao, Deep density estimation via invertible
  block-triangular mapping, Theoretical \& Applied Mechanics Letters 10 (2020)
  143--148.

\bibitem{wan2022vae}
X.~Wan, S.~Wei, {VAE-KR}net and its applications to variational {B}ayes,
  Communications in Computational Physics 31~(4) (2022) 1049--1082.

\bibitem{tang2022adaptive}
K.~Tang, X.~Wan, Q.~Liao, Adaptive deep density approximation for
  {F}okker-{P}lanck equations, Journal of Computational Physics 457 (2022)
  111080.

\bibitem{de2005tutorial}
P.-T. De~Boer, D.~P. Kroese, S.~Mannor, R.~Y. Rubinstein, A tutorial on the
  cross-entropy method, Annals of {O}perations {R}esearch 134~(1) (2005)
  19--67.

\bibitem{rubinstein2013cross}
R.~Y. Rubinstein, D.~P. Kroese, The cross-entropy method: a unified approach to
  combinatorial optimization, Monte-Carlo simulation and machine learning,
  Springer Science \& Business Media, 2013.

\bibitem{fiorin2013using}
G.~Fiorin, M.~L. Klein, J.~H{\'e}nin, Using collective variables to drive
  molecular dynamics simulations, Molecular Physics 111~(22-23) (2013)
  3345--3362.

\bibitem{schrodinger2015pymol}
L.~Schr{\"o}dinger, The {PyMOL Molecular Graphics System, Version} 1.8., (No
  Title) (2015).

\bibitem{alnaes2015fenics}
M.~Aln{\ae}s, J.~Blechta, J.~Hake, A.~Johansson, B.~Kehlet, A.~Logg,
  C.~Richardson, J.~Ring, M.~E. Rognes, G.~N. Wells, The {FE}ni{CS} {P}roject
  {V}ersion 1.5, Archive of Numerical Software 3~(100) (2015).

\bibitem{logg2012automated}
A.~Logg, K.-A. Mardal, G.~Wells, Automated solution of differential equations
  by the finite element method: The {FE}ni{CS} book, Vol.~84, Springer Science
  \& Business Media, 2012.

\bibitem{hartmann2019variational}
C.~Hartmann, O.~Kebiri, L.~Neureither, L.~Richter, Variational approach to rare
  event simulation using least-squares regression, Chaos 29~(6) (2019).

\bibitem{jml23nikolas}
N.~Nüsken, L.~Richter, Interpolating between {BSDE}s and {PINN}s: Deep
  learning for elliptic and parabolic boundary value problems, Journal of
  Machine Learning 2~(1) (2023) 31--64.

\bibitem{vershynin2018high}
R.~Vershynin, High-dimensional probability: An introduction with applications
  in data science, Vol.~47, Cambridge University Press, 2018.

\bibitem{wright2021high}
J.~Wright, Y.~Ma, High-dimensional data analysis with low-dimensional models:
  Principles, Computation, and Applications, Cambridge University Press, 2022.

\bibitem{jo2008charmm}
S.~Jo, T.~Kim, V.~G. Iyer, W.~Im, Charmm-gui: a web-based graphical user
  interface for charmm, Journal of Computational Chemistry 29~(11) (2008)
  1859--1865.

\bibitem{brooks2009charmm}
B.~R. Brooks, C.~L. Brooks~III, A.~D. Mackerell~Jr, L.~Nilsson, R.~J. Petrella,
  B.~Roux, Y.~Won, G.~Archontis, C.~Bartels, S.~Boresch, et~al., Charmm: the
  biomolecular simulation program, Journal of Computational Chemistry 30~(10)
  (2009) 1545--1614.

\bibitem{lee2016charmm}
J.~Lee, X.~Cheng, S.~Jo, A.~D. MacKerell, J.~B. Klauda, W.~Im, Charmm-gui input
  generator for namd, gromacs, amber, openmm, and charmm/openmm simulations
  using the charmm36 additive force field, Biophysical Journal 110~(3) (2016)
  641a.

\bibitem{mcinnes2018umap}
L.~McInnes, J.~Healy, J.~Melville, Umap: Uniform manifold approximation and
  projection for dimension reduction, arXiv preprint arXiv:1802.03426 (2018).

\bibitem{zeng2023BKRnet}
L.~Zeng, X.~Wan, T.~Zhou, Bounded {KR}net and its applications to density
  estimation and approximation, arXiv:2305.09063 (2023).

\bibitem{eastman2017openmm}
P.~Eastman, J.~Swails, J.~D. Chodera, R.~T. McGibbon, Y.~Zhao, K.~A. Beauchamp,
  L.-P. Wang, A.~C. Simmonett, M.~P. Harrigan, C.~D. Stern, et~al., Openmm 7:
  {R}apid development of high performance algorithms for molecular dynamics,
  PLoS Computational Biology 13~(7) (2017) e1005659.

\end{thebibliography}

\newpage
\appendix
\section{Derivation of Variational Formulation}\label{sec_appendix_var}
Let $u = q + \gamma \eta$ be the result of a perturbation $\gamma \eta$ of $q$, where $\gamma$ is small and $\eta$ is a differentiable function. Since $q$ is the minimizer of \eqref{eq_committor_var}, for any $\eta$, we have 
\begin{equation} \label{eq_committor_var_derive}
	\begin{aligned}
		0 &=  \frac{1}{2} \frac{\partial}{ \partial \gamma}|_{\gamma = 0} \int_{\Omega \backslash (A \cup B)} \vert \nabla u(\mb{x}) \vert^2 e^{-\beta V(\mb{x})}  d\mb{x} \\
		& = \int_{\Omega \backslash (A \cup B)} \nabla q(\mb{x}) \cdot \nabla \eta (\mb{x}) e^{-\beta V(\mb{x})} d\mb{x} \\
		& = \int_{\Omega \backslash (A \cup B)} \nabla \cdot \left( \nabla q(\mb{x}) \eta (\mb{x}) e^{-\beta V(\mb{x})}\right) d\mb{x} - \int_{\Omega \backslash (A \cup B)} \eta (\mb{x}) \nabla \cdot \left( \nabla q(\mb{x}) e^{-\beta V(\mb{x})} \right) d\mb{x} \\
		& = - \int_{\Omega \backslash (A \cup B)} \eta (\mb{x}) \nabla \cdot \left( \nabla q(\mb{x}) e^{-\beta V(\mb{x})} \right) d\mb{x}\\
		& = - \int_{\Omega \backslash (A \cup B)} \eta (\mb{x}) e^{-\beta V(\mb{x})}  \left( \Delta q(\mb{x}) - \beta \nabla V(\mb{x}) \cdot \nabla q(\mb{x}) \right)d\mb{x},
	\end{aligned}
\end{equation}
where the fourth equality follows from the integration by parts and the Neumann condition in \eqref{eq_committor_pde}. Because \eqref{eq_committor_var_derive} holds for any $\eta$, we have $ \Delta q(\mb{x}) - \beta \nabla V(\mb{x}) \cdot \nabla q(\mb{x}) = 0$, which is the desired PDE form of the committor function.

\section{Implementation Details}\label{sec_appdix_numdetail}
\subsection{Rugged Mueller Potential}\label{sec_appendix_setting_rmp}
We choose a four-layer fully connected neural network $q_{\mb{\theta}}$ with $100$ neurons to approximate the solution. The activation function is chosen to be the hyperbolic tangent function for the hidden layers and the sigmoid function for the output layer. For KRnet, we take five blocks and eight affine coupling layers in each block. A two-layer fully connected neural network with $120$ neurons is employed in each affine coupling layer. The activation function of KRnet is the rectified linear unit (ReLU) function. To generate points in $\Omega \backslash (A \cup B)$, we use the KRnet to learn the sampling distribution $p_{V, q}(\mb{x})=\vert \nabla q_{\mb{\theta}}(\mb{x}) \vert^2 e^{-\beta V(\mb{x})}$ in the box $\left [ -1.5,1 \right ]\times \left [ -0.5,2 \right ]\times\left [ -1,1 \right ]^{d-2}$, and then remove points within the region $A$ and $B$. This can be done by adding a logistic transformation layer \citep{tang_das} or a new coupling layer proposed in \citep{zeng2023BKRnet}. We set $\lambda = 10$ in \eqref{eq_committor_uncons}. The learning rate for the ADAM optimizer is set to $0.0001$, with a decay rate $0.8$ applied every $200$ epochs for training $q_{\mb{\theta}}$ and no decay for training KRnet, and the batch size is set to $m = m^{\prime} = 5000$. The numbers of adaptivity iterations is set to $N_{\rm{adaptive}} = 30$ when $N_e = N_e^{\prime} = 50$ in Algorithm \ref{alg_dastr}. In this test problem, we replace all the data points in the current training set with new samples. 

It is difficult to sample in the transition state region when simulating the SDE. We implement the artificial temperature method as the baseline. More specifically, we increase the temperature by setting $\beta' = 1/20$ to obtain the modified SDE. This modified Langevin equation is solved by the Euler-Maruyama scheme with the time step $\Delta t = 10^{-5}$. With this setting, the data points are sampled from the trajectory of the modified Langevin equation. In this example, we compare the results obtained from DASTR with those from the artificial temperature method.

\subsection{Standard Brownian Motion}\label{sec_appendix_setting_sbm}
We choose a four-layer fully connected neural network $q_{\mb{\theta}}$ with $100$ neurons to approximate the solution, and the activation function of $q_{\mb{\theta}}$ is set to the square of the hyperbolic tangent function. For KRnet, we take five blocks and eight affine coupling layers in each block. A two-layer fully connected neural network with $120$ neurons is employed in each affine coupling layer. The activation function of KRnet is the rectified linear unit (ReLU) function. The learning rate for the ADAM optimizer is set to $0.001$, with a decay rate $0.8$ applied every $200$ epochs for training $q_{\mb{\theta}}$ and no decay for training KRnet. We set the number of adaptivity iterations to $N_{\rm{adaptive}} = 30$, with $N_e = N_e^{\prime} = 50$ training epochs per stage. The batch size for training $q_{\mb{\theta}}$ is set to $m = 1000$ and for training the PDF model is set to $m^{\prime} = 5000$. In the first stage, we generate $N_0$ uniform samples from \(\Omega \backslash (A \cup B)\) and $N_0 / 2$ points each from \(\partial A\) and \(\partial B\). For the remaining stages, we select $N_0/2$ points from the uniform samples and $N_0/2$ points from the deep generative model. We set $\lambda = 1000$ in \eqref{eq_committor_uncons}.

We use the deep generative model to approximate $p_{V, q}(\mb{x}) = \vert \nabla q_{\mb{\theta}}(\mb{x}) \vert^2 e^{-\beta V(\mb{x})}$, where the probability density function induced by the deep generative model is defined in the box $\left [ -2,2 \right ]^{d}$. To ensure points in $\Omega \backslash (A \cup B) $, we just remove points within the region $A$ and $B$ generated by the deep generative model. For comparison, we also use the SDE to generate data points to train $q_{\mb{\theta}}$, where the Euler-Maruyama scheme with the time step $\Delta t = 10^{-6}$ is applied to get the trajectory.

\subsection{Alanine Dipeptide}\label{sec_appendix_setting_ad}
\paragraph{DASTR with Explicit Collective Variables}
In this test problem, we choose the dihedrals $\phi$ (with respect to C-N-CA-C), $\psi$ (with respect to N-CA-C-N) as the collective variables (CVs). For this realistic example, it is not suitable to use the uniform samples as the initial training set, since uniform samples are not effective for solving this high-dimensional ($d = 66$) problem and also do not obey the molecular configuration. We use metadynamics to generate samples as the initial training set. 

Metadynamics is an enhanced sampling technique to explore free energy landscapes of complex systems. The idea of metadynamics is to add a history-dependent biased potential to the system to discourage it from revisiting previously sampled states \citep{bussi2020using, barducci2008well}. This is done by periodically depositing Gaussian potentials along the trajectory of the CVs. Mathematically, the Gaussian potential can be expressed as:
\begin{equation}\label{Guassian_potential}
	V_{G,t}(\mb{x}) = \sum_{t^{\prime}=0, \tau, 2\tau, \ldots}^{t^{\prime} < t} w \exp \left( -\sum_{i=1}^{m} \frac{(s_i(\mb{x}) - s_i(\mb{x}_{t'}))^2}{2\sigma_i^2} \right), 
\end{equation} 
where $w$ is the height of the Gaussian potential, $\sigma$  is the width of the Gaussian potential, $m$ is number of CVs, and $s_i(\mb{x}_t)$ denotes the collective variables at time $t$. After adding the above Gaussian potential, we generate samples using the modified potential:
\begin{equation*}
	V_{\text{modified}}(\mb{x}) = V(\mb{x}) + V_{G, t}(\mb{x}),
\end{equation*}
where $V(\mb{x})$ is the original potential. That is, the biased potential in \eqref{eq_committor_pdf2} is the Gaussian potential function $V_{G,t}$.
During the simulation, the Gaussian potential lowers the energy barrier, allowing the system to explore more configurations of molecules. So, we can generate effective data points as the initial training set by metadynamics for this alanine dipeptide problem.

\begin{figure}[!ht]
	\centering
	\subfloat[][Samples under the original potential. ]{\includegraphics[width=.35\textwidth]{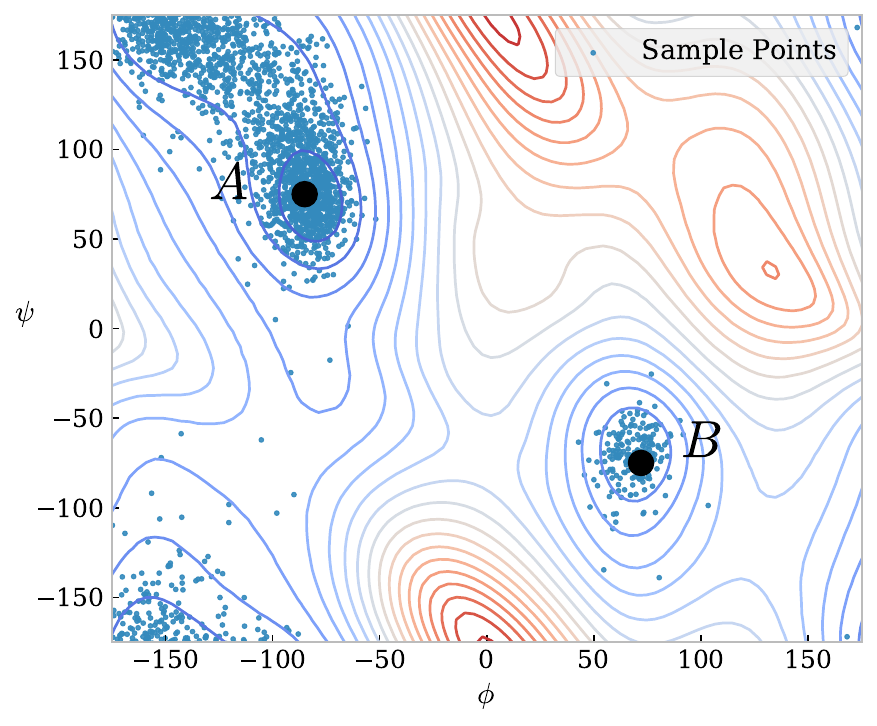}}\quad
	\subfloat[][Samples by metadynamics with 5000 Guassian terms. ]{\includegraphics[width=.35\textwidth]{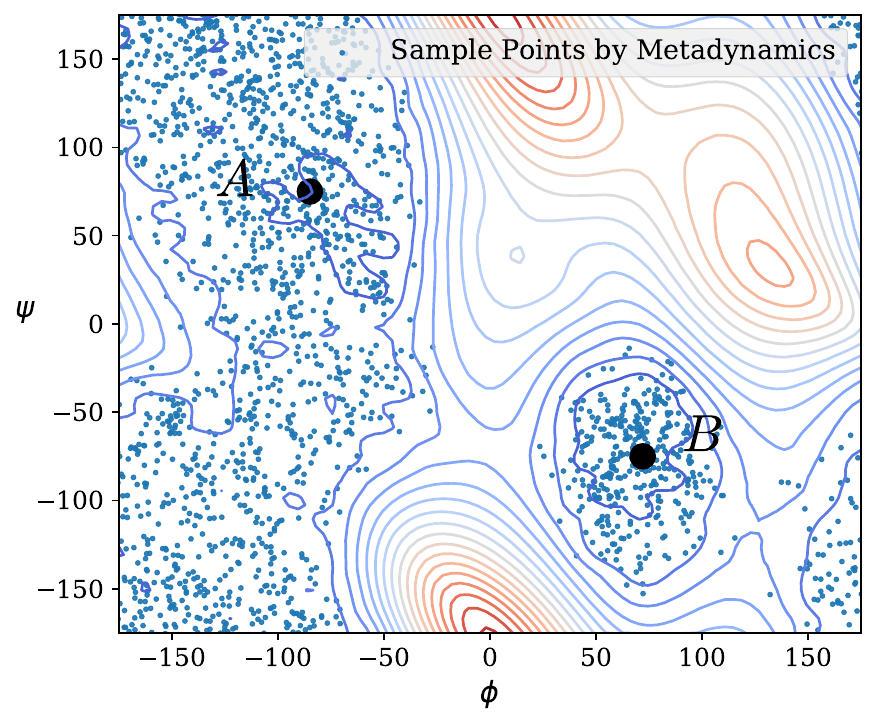}}\\
	
	\subfloat[][Samples by metadynamics with 7500 Guassian terms. ]{\includegraphics[width=.35\textwidth]{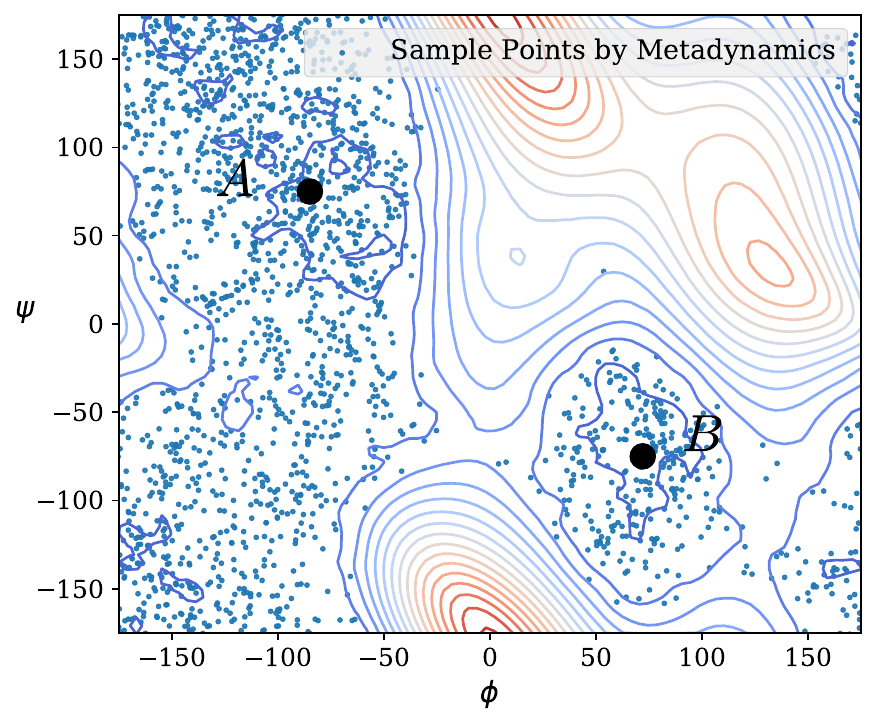}}\quad
	\subfloat[][Samples by metadynamics with 10000 Guassian terms. ]{\includegraphics[width=.35\textwidth]{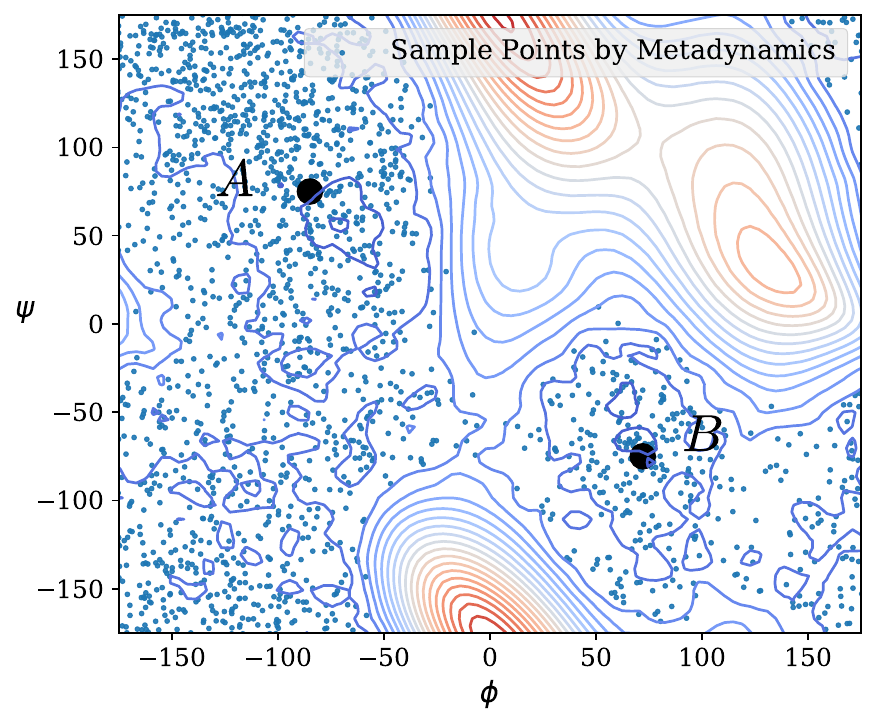}}
	\caption{Samples from the original dynamics and metadynamics. }
	\label{fig:ad_sample_meta}
\end{figure}

We simulate the Langevin dynamics with the time step $\Delta t = 0.1 \, \text{fs}$ and a damping coefficient $1 \, \text{ps}^{-1}$. One term of the Gaussian potential is added every $1000$ steps, with parameters $w = 1.0 \, \text{kJ/mol}$, $\sigma = 0.1 \, \text{rad}$. We finally get a total of $5000$ terms in \eqref{Guassian_potential}. Then we conduct the Metadynamics with 7500 and 10000 terms for comparison. Figure~\ref{fig:ad_sample_meta} shows that the more terms we add, the more thoroughly the free energy surface is explored, and the more samples we obtain in the transition state region. Samples are selected outside the regions $A$ and $B$, and system configurations are saved to conduct the importance sampling step in \eqref{eqn_update_theta}. The simulation is conducted in OpenMM \citep{eastman2017openmm}, a molecular dynamics simulation toolkit with high-performance implementation. 
Figure~\ref{fig:ad_sample_meta} shows the samples from the original dynamics and metadynamics. From this figure, it is clear that using metadynamics to generate initial data points is better since more samples are located in the transition state region. 

We choose a five-layer fully connected neural network $q_{\mb{\theta}}$ (with $100, 120, 150$ neurons) to approximate the solution, and the activation function for the  hidden layers is set to the hyperbolic tangent function. The activation function for the output layer is the sigmoid function. Here, we only use the deep generative model to model the sampling distribution in terms of the collective variables $\phi$ and $\psi$. The trained KRnet is used to generate $s(\mb{x}_0) = [\phi, \psi]^\top$ in \eqref{Umbrella_potential} (see \ref{sec_appendix_us}). For KRnet, we take one block and six affine coupling layers in each block. A two-layer fully connected neural network with $64$ neurons is employed in each affine coupling layer. The activation function of KRnet is the rectified linear unit (ReLU) function. The learning rate for the ADAM optimizer is set to $0.0001$, with a decay factor of 0.5 applied every 200 epochs for training $q_{\mb{\theta}}$ and no decay for training KRnet.  We set the batch size $m = 5000, m^{\prime} = 10000$ and $N_e = 300, N_e^{\prime} = 1000$. The numbers of adaptivity iterations is set to $N_{\rm{adaptive}} = 10$. We sample $1.5 \times 10^{4}$ points in $A$ and $B$ respectively to enforce the boundary condition in the training process for all stages. We set $\lambda = 10$ in \eqref{eq_committor_uncons}.

We employ KRnet to learn the sampling distribution in \eqref{eq_committor_pdf2}. In the first stage, we train the neural network $q_{\mb{\theta}}$ using $2 \times 10^5$ points sampled by metadynamics. Then we use these points to train the PDF model induced by KRnet with support $\left [ -180^{\circ},180^{\circ} \right ]^{2}$, with the bias potential $V_{\text{bias}}$ in \eqref{eq_committor_pdf2} being the Gaussian potential $V_{G, t}$ defined in \eqref{Guassian_potential}. In the rest of the stages, we train the neural network $q_{\mb{\theta}}$ with $5 \times 10^4$ points sampled by umbrella sampling with the bias potential $V_{\text{US}}$ (see \ref{sec_appendix_us}). We train the KRnet using the same sample points as those of training $q_{\mb{\theta}}$. 

During the training procedure, we increase $k_{\text{us}}$ in \eqref{Umbrella_potential} from $200$ kJ/mol to $400$ kJ/mol. We sample $100$ points for each target CVs in the umbrella sampling procedure. For comparison, we use the solution obtained by training a neural network $q_{\mb{\theta}}$ with $150$ neurons with $2 \times 10^{5}$  points sampled via metadynamics for $3000$ epochs.

\paragraph{DASTR with Latent Collective Variables}
In this experiment, both the encoder and decoder are implemented using fully connected neural networks. The encoder architecture is set as $[30, 100, 50, 50, 30, d_{\text{latent}}]$, while the decoder is set as $[d_{\text{latent}}, 30, 50, 50, 100, 30]$, with the Swish activation function. For training the autoencoder, we use $2\times 10^5$ samples generated by metadynamics (with 10000 terms in \eqref{Guassian_potential}) as the training set. The batch size is set to 1000. The model is trained with 5000 epochs. 

The committor function is approximated by a five-layer fully connected neural network $q_{\mb{\theta}}$ with 150 neurons, where the activation function for the hidden layers is set to the hyperbolic tangent function, and the activation function for the output layer is the sigmoid function. In this experiment, we use the deep generative model to model the probability distribution in terms of the latent CVs obtained from the autoencoder. The learning rate for the ADAM optimizer is set to 0.0001, with a decay factor of 0.5 applied every 200 epochs for training $q_{\mb{\theta}}$ and no decay for training KRnet. The batch size is set to $m = 5000, m^{\prime} = 10000$ and $N_e = 200, N_e^{\prime} = 500$. In the first stage, we use \( 2 \times 10^5 \) points sampled from metadynamics (10000 terms in \eqref{Guassian_potential}) as the initial dataset to train $q_{\mb{\theta}}$. In the rest stages, we use \( 1 \times 10^5 \) points sampled from metadynamics and \( 1 \times 10^5 \) points from KRnet and the pretrained autoencoder. Other settings are the same as those in section \ref{sec_numexp_ad_explicit_CVs}.

\subsection{Umbrella Sampling}\label{sec_appendix_us}
The umbrella sampling method is also an enhanced sampling technique. It introduces external biased potentials to pull the system out of local minima, thereby enabling a more uniform exploration of the entire free energy surface. This method is particularly effective in calculating free energy differences and studying reaction pathways in complex molecular processes. The umbrella sampling method employs a series of biased simulations, dividing the reaction space of collective variables into multiple overlapping windows, where each biased potential is applied in its corresponding window \citep{kastner2011umbrella}.
The umbrella potential is usually defined as:
\begin{equation}\label{Umbrella_potential_appd}
	V_{\text{US}}(\mb{x}) = \frac{1}{2} \sum_{i=1}^{m} k_{\text{us}} (s_{i}(\mb{x}) - s_{i}(\mb{x}_0))^2, 
\end{equation}
where $s_{i}(\mb{x})$ represents the CVs with respect to $\mb{x}$, $m$ is the number of CVs, and $k_{\text{us}}$ is the force constant. 
In this work, we focus on sampling in the final window, helping us effectively sample the desired regions of CVs. Therefore, we perform a rapid iterative process of umbrella sampling to transfer the CVs to the target region, and finally sample near the target CVs in the modified potential:
\begin{equation*}
	V_{\text{modified}}(\mb{x}) = V(\mb{x}) + V_{\text{US}}(\mb{x}),
\end{equation*}
where $V$ is the original potential, and $s_{i}(\mb{x}_0)$ in \eqref{Umbrella_potential_appd} is the target CVs generated by the trained deep generative model.

\end{document}